\documentclass[conference]{IEEEtran}

\usepackage{cite}
\usepackage{amsmath,amssymb,amsfonts}
\usepackage{algorithmic}
\usepackage{graphicx}
\usepackage[normalem]{ulem}
\usepackage{textcomp}
\usepackage{xcolor}
\usepackage[margin=1pt]{subfig}
\usepackage{multirow}
\def\BibTeX{{\rm B\kern-.05em{\sc i\kern-.025em b}\kern-.08em
    T\kern-.1667em\lower.7ex\hbox{E}\kern-.125emX}}

\DeclareRobustCommand*{\IEEEauthorrefmark}[1]{%
  \raisebox{0pt}[0pt][0pt]{\textsuperscript{\footnotesize #1}}%
}

\begin{document}

\title{PatchX: Explaining Deep Models by Intelligible Pattern Patches for Time-series Classification.}

\author{\IEEEauthorblockN{Dominique Mercier\IEEEauthorrefmark{1}\IEEEauthorrefmark{2}, Andreas Dengel\IEEEauthorrefmark{1}\IEEEauthorrefmark{2}, Sheraz Ahmed\IEEEauthorrefmark{2}}
\IEEEauthorblockA{\IEEEauthorrefmark{1}Technische Universität Kaiserslautern, 67663 Kaiserslautern, Germany.}
\IEEEauthorblockA{\IEEEauthorrefmark{2}German Research Center for Artificial Intelligence (DFKI), 67663 Kaiserslautern, Germany.}
\IEEEauthorblockA{EMail: firstname.lastname@dfki.de}}

\maketitle

\begin{abstract}
The classification of time-series data is pivotal for streaming data and comes with many challenges. Although the amount of publicly available datasets increases rapidly, deep neural models are only exploited in a few areas. Traditional methods are still used very often compared to deep neural models. These methods get preferred in safety-critical, financial, or medical fields because of their interpretable results. However, their performance and scale-ability are limited, and finding suitable explanations for time-series classification tasks is challenging due to the concepts hidden in the numerical time-series data. Visualizing complete time-series results in a cognitive overload concerning our perception and leads to confusion. Therefore, we believe that patch-wise processing of the data results in a more interpretable representation. We propose a novel hybrid approach that utilizes deep neural networks and traditional machine learning algorithms to introduce an interpretable and scale-able time-series classification approach. Our method first performs a fine-grained classification for the patches followed by sample level classification.
\end{abstract}


\maketitle

\section{Introduction}
Besides image classification and natural language processing, the time-series classification is one of the most challenging problems in data mining~\cite{yang200610, esling2012time}. Although the amount of publicly available time-series data~\cite{silva2018speeding, dau2019ucr}, and time-series approaches~\cite{bagnall2017great} are increasing continuously, the exploitation of deep models on these growing datasets is still limited. In theory, any ordered data can be considered as a time-series classification problem~\cite{langkvist2014review, gamboa2017deep} leading to the impressive amount of available time-series data. When used in the medical~\cite{rajkomar1801scalable}, financial domain or in safety-critical~\cite{susto2018time} applications besides the performance it is equally or sometimes even more important to understand the reason for particular decisions made by the system. Therefore the XAI aims to find suitable explanations for the model inference process and the data~\cite{mercier2020interpreting} to make to provide trust on a user-level.

Model results can be explained from multiple perspectives e.g., data, model insights, etc. Most intuitive explanations are those where one can relate a classified sample to another known sample of the same class based on its similarity. For instance, an image is classified as a car because it shares some concepts with a reference image of a car. In the image domain, these kinds of concepts~\cite{nixon2019feature} are well defined e.g. wheels, car windows, etc. Similar this is the case for the natural language processing~\cite{collobert2011natural}. However, concept identification is quite difficult in time-series because the concepts in time-series data are not directly intelligible~\cite{christ2018time}, which makes it harder to both annotate and detect these concepts in time-series data. To overcome this problem, this paper presents a novel approach that performs a piece-wise fine-grained classification of time-series. The explanation covers a visually perceivable amount of data~\cite{yang2018learning} corresponding to concepts~\cite{dastani2002role} to lower the cognitive load.

Our approach reduces the complexity of the problem by implementing a divide and conquer principle to reduce the cognitive load required to understand a prediction. For each patch, we compute its relevance for the different classes. We believe that the possible unlimited amount of time-steps and channels is one source of complexity, and it is beneficial to explain smaller subsets of the data. Therefore, we discover the features and class relation for the separated patches. Later on, we combine the results of the patch-wise processing to achieve an overall classification ~\cite{geurts2001pattern, xing2011extracting, ghalwash2013extraction}.

Our proposed approach, i.e., PatchX performs a fine-grained patch classification using deep learning methods followed by overall sample classification with a traditional machine learning classifier. Therefore, we can interpret the fine-grained patches to understand the individual fine-grained blocks and their influence on the overall classification. We further demonstrate on multiple publicly available datasets that our approach can compete with other existing approaches concerning the accuracy, interpretability, and scalability capabilities and is not limited to the time-series domain. Our modular design allows exchanging the network and the classifier based on the needs. Our four contributions are the following:

\begin{itemize}
    \item A novel data transformation approach for time-series that creates patches of different sizes and transforms them into a shape usable by every neural network.
    \item An interpretable hybrid approach for time-series classification based on local patterns.
    \item An extensive evaluation of existing approaches along with the PatchX on five publicly available datasets. The evaluation results show the superiority of PatchX in comparison to SotA with an added benefit of interpretability. 
\end{itemize}

\begin{table}[!t]
\renewcommand{\arraystretch}{1.3}
\caption{\textbf{Interpretability approaches.} Different existing interpretability approaches applicable to time-series.}
\label{tab:related}
\centering
\scalebox{0.72}{
\begin{tabular}{|c|c|c|c|c|}
    \hline
	\textbf{Paper} & \textbf{Explanation} & \textbf{Expl. scope} & \textbf{Data scope} & \textbf{Domain} \\
	\hline
	Xing et al.~\cite{xing2011extracting} & pattern & local & instance & time-series \\
	Ye et al.~\cite{ye2011time} & pattern & local & instance & time-series \\
	Mercier et al.~\cite{mercier2020p2exnet} & prototypes & local & global & time-series \\
	Siddiqui et al.~\cite{siddiqui2019tsviz} & saliency & global& instance & time-series \\
	Schlegel et al.~\cite{schlegel2019towards} & LRP & global& instance & time-series \\
	Siddiqui et al.~\cite{siddiqui2020tsinsight} & compression & global & instance & time-series \\
	Millan and Achard\cite{millan2020explaining} & saliency & global & instance & time-series \\
	Gee et al.~\cite{gee2019explaining} & prototypes & global & global & time-series \\
	\hline
	\hline
	Frid-Adar et al.~\cite{frid2017modeling} & patches & local & instance & images \\
	Zhang et al.~\cite{zhang2018interpretable} & pattern & local & instance & images \\
	Fong and Vedaldi\cite{fong2018net2vec} & pattern & local & instance & images \\
	Hou et al.~\cite{hou2016patch} & patches & local \& global & instance & images \\
	Selvaraju et al.~\cite{selvaraju2017grad} & saliency & global & instance & images\\
	Kim and Canny\cite{kim2017interpretable} & saliency & global & instance & images \\
	Schneider and Vlachos\cite{schneider2020explaining} & prototype & global & global & images \\
	\hline
	\hline
	Ribeiro et al.~\cite{ribeiro2016should} (LIME) & patches & local & instance & tabular \\
	Lundberg et al.~\cite{lundberg2017unified} (SHAP) & saliency & local & instance \& global & tabular \\
	\hline
	\hline
	Ours & pattern & local \& global & instance \& global & time-series \\
	\hline
\end{tabular}
}
\end{table}

\section{Related Work}
To evaluate our contribution it is important to understand the existing methods and the perspectives used to interpret the model. Therefore, we cover interpretability methods concerning patch-wise processing, saliency methods, and heat-map approaches. All these methods try to identify the features used by the model. Although we are focusing on time-series data, we mention approaches designed for the image domain as these can be transferred to the time-series domain or their perspective of the problem statement, and their solution is related to ours. We list these methods in Table~\ref{tab:related} to compare their focus. As our method can provide local as well as global explanations for a given instance or in a global manner for a class, the related work covers approaches related to any of these properties. Therefore, we explain and summarize the most relevant aspects in the following sections.

\subsection{Explanation Scope: Local Regions or Complete Sample}
A local explanation scope explains parts of the data making it possible to produce instance-based explanations. The extraction of patches~\cite{ribeiro2016should, frid2017modeling, hou2016patch} and the definition of patterns~\cite{xing2011extracting, ye2011time, zhang2018interpretable, fong2018net2vec} are most times used to produce these explanations. Especially, the patch-based approach is widespread in the image domain as the modalities allow to directly interpret the segmented parts of the images. Although this approach can be transferred to the time-series domain, it results in a more complex interpretation task. In contrast to the patches, pattern extraction aims to go one step further and creates abstract patterns that are similar to the original patches. This abstraction removes noise and unrelated differences providing a more understandable pattern. However, they only provide a shallow relation to the original samples.

Another common approach is the global explanation scope enabling the interpretation of the complete sample data at once. Utilizing the computation of the gradients~\cite{siddiqui2019tsviz, millan2020explaining, selvaraju2017grad, kim2017interpretable, lundberg2017unified} it is possible to compute saliency maps and get an importance value for each point in the input signal. Also, it is possible to compute these importance values for different parts of the network and try to understand the outcome of the corresponding neurons. There exist many different related approaches aiming to improve the robustness against adversarial attacks and noise~\cite{melis2018towards}. However, these methods do not include information about the concept behind the highlighted points. An approach that is related to the saliency methods is the layer-wise relevance propagation~\cite{schlegel2019towards} as it computes the relevant parts of the input and creates a heat-map similar to saliency methods. All these methods have shown great success in the image domain as the highlighting enables humans to interpret them and extract the underlying concept themselves.

\subsection{Data Scope: Instance-based vs Global}
The data scope describes whether the explanation is instance-based or covers global behavior. Usually, a global explanation is based on behavior that holds across the instances. Global methods are mostly prototype-based~\cite{mercier2020p2exnet, gee2019explaining, schneider2020explaining}. An advantage of these methods compared to the instance-based methods is their abstraction and generalization. The definition of class prototypes leads to understandable prototypes but requires a careful selection and a good comparison algorithm. Furthermore, these approaches providing a set of prototypes for different parts of a sample can lead to a comprehensive explanation~\cite{mercier2020p2exnet}. However, global data scope methods do not take into account edge cases that can be of high importance, especially in a safety-critical task or error handling. 

On the other side, instance-based explanations specific to the input instance are widespread as their complexity is much lower. Furthermore, an instance-based explanation is closely aligned with the given input making it much easier to detect misbehavior. 

In conclusion, both the local and global explanations have advantages. The local explanations are more related to simple and concept-based approaches but do not capture the global behavior of the network. The global prototypes may be difficult to explain as they only cover the global behavior. 

\subsection{Literature Summary}
All methods mentioned in Table~\ref{tab:related} can be adapted to work with time-series data and provide different explanations using different perspectives. When it comes to their effectiveness there is no clear winner as it is difficult to compare them concerning the different scopes. Finally, for further information about the advantages and drawbacks of the different methods we refer to~\cite{zhang2018visual} for image related approaches and~\cite{mohseni2018survey} for an overall survey of interpretability. Both papers provide a summary of the most famous approaches and the different perspectives used by those.

\section{Method}

\begin{figure}[!t]
\centering
\includegraphics[width=1.0\linewidth]{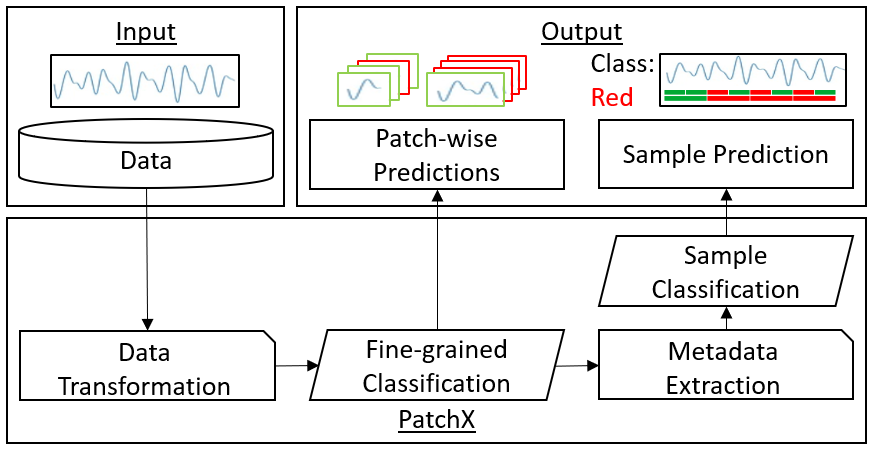}
\caption{\textbf{Workflow.} PatchX only requires the time-series to produce a fine-grained and a overall prediction.}
\label{fig:pipeline}
\end{figure}

We propose a hybrid approach using a neural network and a traditional machine learning approach to infer on the patch and the sample level. Our architecture is modular and both the network or the classifier can be exchanged. Our approach works with automatically created metadata for the patches created during the processing. In the following sections, we provide detailed information about the automatic label creation, required data transformation steps, the classification process, and the mathematical background. The processing of a sample is divided into four steps shown in Figure~\ref{fig:pipeline}.

\subsection{Data Transformation (Step 1)}
The initial processing step transforms the sample into patches used for further processing. To do so, a set of patch sizes and strides is passed to the framework that is used to define the boundaries. This set can include a guess as the framework automatically handles the relevance of the provides sets. During the data transformation, every patch gets the label of its corresponding sample. An important aspect is that the transformation preserves the length of the data enabling us to use different patch sizes within the same network. To do so, the data not related to the patch is set to zero, and an additional channel highlighting the true length of the patch is created. 

Let $x_{i}^{p}$ denote the transformed sample for the \textit{p}-th patch of the \textit{i}-th sample in \textit{X}. Equation~\ref{eq:x_patch} shows the required parameters to compute $x_{i}^{p}$ using the index \textit{p} of the patch, \textit{s} as the given stride, and \textit{l} as the patch size. In Equation~\ref{eq:x_patches}, we show the computation of $\hat{X}$ the set of all patches over each sample. The transformed data $\hat{X}$ consisting of the different patches for each sample and the label $y_{i}$ corresponding to the $x_{i}$ is used to create $x_{i}^{p}$. 

\begin{equation}
x^{p}_{i} = transform(x_{i}, p, s, l)
\label{eq:x_patch}
\end{equation}

\begin{equation}
\hat{X} = \left \{ x_{i}^{p} \mid i \in N \wedge p \in \mathbb{N} \wedge p * s < \mid x_{i} \mid \right \}
\label{eq:x_patches}
\end{equation}

Due to simplicity reasons, we assume that only a single pair of \textit{s} and \textit{l} is used. However, the equations do not change dramatically as only the number of patches per sample increases, and a transformation for each setup is applied. 

\subsection{Fine-grained Classification (Step 2)}
In the second step, the patch data is used to train a deep neural network to perform a fine-grained classification on the patch level. Although the dataset contains only the overall labels, it is possible to learn a patch-based behavior. The minimization of the network loss focuses on the samples that are class-specific. Therefore, the softmax prediction shows uncertainty for the patches that appear in different classes. This uncertainty highlights that the patches are not class-specific and shared between classes serving as a confidence score.  

Equation~\ref{eq:h_patch} shows the cross-entropy for a single patch. In Equation~\ref{eq:h_patches}, the loss over all samples is visualized. Therefore, we denote \textit{P} as the set of patches per sample and \textit{N} as dataset size corresponding to the size before splitting the data into patches. Furthermore, we denote the patch classification network as $\Phi$ and \textit{C} as the classes.

\begin{equation}
H = \sum_{c=1}^{C} -y_{i,c} * log( \Phi_{c} (x_{i}^{p}))
\label{eq:h_patch}
\end{equation}

\begin{equation}
L = \frac{1}{\mid N \mid * \mid P \mid } * \sum_{i}^{N} \sum_{p}^{P} H(x_{i}^{p}, y_{i})
\label{eq:h_patches}
\end{equation}

\textbf{Discussion:} Piece-wise processed data is not available most time and its manual annotation is not suitable. Therefore, assigning the sample label to a patch introduces the question of whether it is correct or not. In general, each patch can be classified as part of one of the following categories~\cite{kim2018interpretability}. 

\begin{itemize}
    \item The pattern occurs only in a single class. In this case, the label is correct, and the classification of the patch will result in a very high value for the assigned class. We refer to this pattern as a class-specific pattern.
    \item The pattern occurs in multiple classes. Based on the label creation process, the same pattern has different labels. The network prediction for these samples will have medium values for some classes. These patterns are referred to as shared patterns.
    \item The pattern is not related to the label. These patches are actual mislabels, but due to the minority of this case, the loss minimization can handle them.
\end{itemize}

\subsection{Metadata Extraction (Step 3)}
The softmax prediction given by the fine-grained classification is then used to create a suitable representation for the sample classification. Therefore, a vector with the shape of the different classes is initialized. This vector is filled with the softmax values using only the maximum value of each patch prediction and adding it to the corresponding class index. The vector is time-independent but represents the presence of different classes within the sample.

In Equation~\ref{eq:x_hist_c} we show the computation of the entry $\tilde{x}_{i}^{c}$. We denote $\tilde{x}_{i}$ as the feature vector used to compute $x_{i}$. The values of each $\tilde{x}_{i}^{c}$ are represented by the sum of the values corresponding to the class \textit{c} of the patches for that \textit{c} was predicted. In Equation~\ref{eq:x_histo} we show $\tilde{x}_{i}$ as set of $\tilde{x}_{i}^{c}$ computed for each class $c$ in $\mid C \mid$ the set of classes.

\begin{equation}
\tilde{x}_{i}^{c} = \sum_{p}^{P} \left \{ max(\Phi (x_{i}^{p})) \mid argmax(\Phi (x_{i}^{p})) = c \right \}
\label{eq:x_hist_c}
\end{equation}

\begin{equation}
\tilde{x}_{i} = \left \{ \tilde{x}_{i}^{c} \mid c \in C \right \}
\label{eq:x_histo}
\end{equation}

\textbf{Discussion:} The drop of the temporal component during the metadata extraction can be explained with the previous patch prediction. The patch prediction is executed in a manner that takes care of the temporal component and inherently encodes the temporal component. This makes it possible to simply add the values to the corresponding class. Separating the class vector to preserve the patch sequence or the use of the complete metadata resulted in no information gain.

\subsection{Sample Classification (Step 4)}
The last step is the classification based on the prepared metadata using a traditional machine learning algorithm. We decided to go with a support vector machine or random forest classifier as these had the best performance during our experiments. However, it is possible to replace them with any other classifier e.g. a dense layer, nearest neighbor, or majority/occurrence voting.

In Equation~\ref{eq:y_clf} the final predictions are computed using the $\tilde{x}$ as feature vector created for each $x_{i}$ of the dataset. 

\begin{equation}
{y}' = \left \{ \Psi (\tilde{x}_{i}) \mid i \in N \right \}
\label{eq:y_clf}
\end{equation}

\section{Datasets}
We used a set of five publicly available time-series datasets to show that our approach works without a restriction to a specific dataset. Precisely, we used four datasets from the UCR Time-Series Classification Repository~\footnote{http://www.timeseriesclassification.com/}, and a point anomaly dataset proposed in~\cite{siddiqui2019tsviz}. We selected these sets to emphasize broad applicability and possible limitations. In Table~\ref{tab:datasets} we show the different parameters of each dataset. During the dataset selection, we selected datasets that differ in the number of channels, time-steps, classes, dataset size, and task.

\begin{table}[!t]
\caption{\textbf{Dataset parameter}. The different datasets and their parameters.}
\label{tab:datasets}
\centering
\scalebox{0.8}{
\begin{tabular}{|c|c|c|c|c|c|c|}
\hline
\multirow{2}{*}{\textbf{Dataset}} & \multicolumn{3}{c|}{\textbf{Instances}} & \multirow{2}{*}{\textbf{Classes}} & \multirow{2}{*}{\textbf{Length}} & \multirow{2}{*}{\textbf{Channel}} \\
& \textbf{Train} & \textbf{Val} & \textbf{Test} & & & \\
\hline
ElectricDevices & 6244 & 2682 & 7711 & 7 & 96 & 1 \\
DailyAndSportActivities & 22344 & 9576 & 13680 & 19 & 60 & 45 \\
CharacterTrajectories & 1383 & 606 & 869 & 20 & 206 & 3 \\
FordA  & 2520 & 1081 & 1320 & 2 & 500 & 1 \\
Anomaly~\cite{siddiqui2019tsviz} & 35000 & 15000 & 10000 & 2 & 50 & 3 \\
\hline
\end{tabular}
}
\end{table}

\section{Experiments and Results}
A high-quality interpretable network architecture requires not only good accuracy but further requires a suitable explanation and broad applicability. Therefore, we performed several experiments including a comparison of non-interpretable, and interpretable approaches, and different accuracy evaluations. These experiments highlight the performance, show different explanations and use cases. 

\subsection{Accuracy Comparison}

\begin{table*}[!t]
\renewcommand{\arraystretch}{1.3}
\caption{\textbf{Accuracy comparison.} \textit{Feature} approaches include a feature extraction preprocessing. The \textit{'Trivial'} approach covers a majority / occurrence voting after the fine-grained classification.}
\label{tab:accuracies_comparison}
\centering
\scalebox{1.0}{
\begin{tabular}{|c|c||c|c|c||c|c|c|c|}
    \hline
    \multirow{4}{*}{\textbf{Dataset}} & \multicolumn{4}{c||}{\textbf{Scalable}} & \multicolumn{4}{c|}{\textbf{Not scalable}} \\
    \cline{2-9}
	& \multirow{2}{*}{\textbf{Blackbox}} & \multicolumn{7}{c|}{\textbf{Interpretable}} \\
	\cline{3-9}
	& & \multicolumn{3}{c||}{\textbf{PatchX}} & \multicolumn{4}{c|}{\textbf{Traditional approaches}}\\
	\cline{2-9}
	& \textbf{CNN} & \textbf{CNN + SVM} & \textbf{CNN + RF} & \textbf{CNN + Trivial} & \textbf{SVM} & \textbf{RF} & \textbf{SVM feature} & \textbf{RF feature} \\
	\hline
	\hline
	ElectricDevices & 67.31 & \textbf{69.29} & 68.56 & 60.68 & 60.69 & 65.21 & 24.23 & 69.46 \\
    \hline
	DailyAndSportActivites & 99.74 & 99.82 &\textbf{ 99.88} & 99.79 & 98.54 & 99.60 & - & - \\
	\hline
	CharacterTrajectories & 96.55 & 95.17 & 94.13 & 82.74 & \textbf{98.62} & 98.16 & 92.98 & 98.16 \\
	\hline
	FordA & 88.71 & \textbf{90.08} & 82.42 & 88.48 & 83.33 & 74.92 & 92.88 & 1.0 \\
	\hline
	Anomaly & \textbf{98.70} & 98.41 & 97.25 & 98.23 & 97.74 & 96.34 & 99.08 & 99.99 \\
	\hline
\end{tabular}
}
\end{table*}

We performed a comprehensive accuracy evaluation over different classification approaches to compare their performances and scale-ability. Besides, we divided the approaches into non-interpretable deep learning, and interpretable methods. Furthermore, within the class of interpretable approaches, we differ between traditional machine learning algorithms and our interpretable hybrid approach. Finally, we included a feature extraction preprocessing step for some methods to evaluate the impact of extracted features against raw data usage. The different approaches and performances are shown in Table~\ref{tab:accuracies_comparison}.

Although deep learning is known to perform well, the results of traditional machine learning including a feature extraction step have shown superior performance. However, this feature extraction approach limits the approaches significantly. E.g. for the \textit{DailyAndSportActivities} dataset the time consumption and memory usage scale very poor making it impossible to use the feature extraction for larger datasets. Besides, the feature extraction assumes that you already know the numerical features of interest and does not learn others directly from the data.

Excluding the limited feature extraction based approaches, deep learning outperforms the traditional approaches in almost every task as the feature extraction is not scale-able. Compared to the traditional approaches, the deep learning algorithms do not suffer from the increased dataset size, and CNNs have shown to be able to work with the raw data and produce high accuracy results. However, the results are not interpretable without additional efforts.

The results show that our hybrid approach is a good compromise as its accuracy is only slightly lower compared to the deep learning approach but it scales very well and produces explainable results. Furthermore, the computation time of the hybrid approach is only slightly higher than the computation of the blackbox approach.

\subsection{Computation Time Analysis}
We experimented to evaluate the scale-ability time consumption of the different approaches. In Figure~\ref{tab:time} we show the results for the complete training procedure and the testing using the test datasets. The results show that the approaches that use the feature extraction scale pretty bad. Conversely, the deep learning-based methods scale pretty well. Furthermore, PatchX is slower than the traditional approach when the dataset is small but scales significantly better when the dataset size increases.

\begin{table}[!t]
\renewcommand{\arraystretch}{1.3}
\caption{\textbf{Time consumption.} \textit{T} denotes to the training time in seconds using a . \textit{I} denotes the inference time over the test dataset. Used hardware: Intel Xeon (Quad Core), Nvidia GTX 1080 Ti, 64 GB memory.}
\label{tab:time}
\centering
\scalebox{0.8}{
\begin{tabular}{|c|c|c|c|c|c|}
    \hline
	\textbf{Dataset} & \textbf{Mode} & \textbf{PatchX}  & \textbf{Blackbox} & \textbf{SVM} & \textbf{SVM Feature} \\
	\hline
	\multirow{2}{*}{ElectricDevices} & T & 83.2 & 7.4  & 9.3 & 161.9 \\
	& I & 4.8 & 0.6 & 5.8 & 12.1 \\
	\hline
	\multirow{2}{*}{DailyAndSportActivities} & T & 220.9 & 32.1 & 511.8 & - \\
	& I & 5.8 & 1.2 & 504.5 & - \\
	\hline
    \multirow{2}{*}{CharacterTrajectories} & T & 69.9 & 6.7 & 0.8 & 132.2 \\
    & I & 1.4 & 0.2 & 0.5 & 45.2 \\
    \hline
    \multirow{2}{*}{FordA} & T & 159.2 & 10.5 & 5.9 & 293.0 \\
    & I & 3.0 & 0.2 & 2.0 & 100.8 \\
    \hline
    \multirow{2}{*}{Anomaly} & T & 168.5 & 45.3 & 1290.0 & 3006.5 \\
    & I & 2.6 & 0.6 & 33.7 & 260.3 \\
    \hline
\end{tabular}
}
\end{table}

\subsection{Hyper-parameter Selection}
To produce these interpretable results a careful hyper-parameter selection is important. Our proposed approach requires two types of parameters. 

\subsubsection{Patch creation parameters}
The first category includes the parameters that directly influence the patches. Precisely, \textit{stride} and \textit{length} are used to define the patches. As in any other use case, the \textit{stride} defines the gap between the patches of the same sample, and the \textit{length} the length of the data. Using our approach it is possible to use multiple \textit{strides} and \textit{lengths}. This results in a larger dataset and different levels of explanation. 

In Table~\ref{tab:accuracies} we show the impact of the different setups and compare them. Therefore, we defined stride to be half of the length to produce an overlap of the samples and used an SVM classifier for the sample classification. The results show that for all datasets except the \textit{FordA} the small patch length was able to capture the pattern required to classify the samples. However, using the increased patch size we were able to recover the network for the \textit{FordA} dataset.

\begin{table}[!t]
\renewcommand{\arraystretch}{1.3}
\caption{\textbf{Stride and length selection.} Performance comparison using different parameter sets. 'S' denotes the stride between the patches and 'L' denotes the length of each patch.}
\label{tab:accuracies}
\centering
\begin{tabular}{|c|c|c|c|}
    \hline
	\multirow{2}{*}{\textbf{Dataset}} & \multicolumn{3}{c|}{\textbf{PatchX}} \\
	& \textbf{S 5 L 10} & \textbf{S 10 L 20} & \textbf{S 5,10 L 10,20} \\
	\hline
	ElectricDevices & 65.78 & 69.06 & \textbf{69.29} \\
    \hline
	DailyAndSportActivites & 99.66 & 99.74 & \textbf{99.82} \\
	\hline
	CharacterTrajectories & \textit{94.82} & 95.17 & 94.36 \\
	\hline
	FordA & 51.59 & \textbf{89.17} & 87.20 \\
	\hline
	Anomaly & 98.27 & 98.14 & \textit{98.29} \\
    \hline
\end{tabular}
\end{table}

The results highlight how crucial the parameter selection is for interpretation and accuracy. Intuitively, smaller patch sizes are related to basic patterns and larger patch sizes cover complex patterns~\cite{li2017linear, feremans2019pattern}. Furthermore, the experiment shows that the use of multiple parameter setups including the not working setup for the \textit{FordA} data resulted in high accuracy values as our approach automatically takes care of setups that are irrelevant for the sample classification. However, it is possible to include those setups as the extracted patches can be used to draw conclusions about the data and create synthetic samples.

\subsubsection{Patch transformation parameters}
Taking into account the variable length of the patches within the experiment settings requires an advanced transformation. Three parameters are required to handle the transformation. Although it is possible to have different combinations of these parameters, only a limited set is valid. 

\begin{enumerate}
    \item \textit{Zero} is used to set the data not included in the patch to zero. As the model has a fixed input size it is mandatory to maintain the sample size for all patch sizes. 
    \item \textit{Attach} indicates the data related to the patch using an additional channel.
    \item \textit{Notemp} removes the time component and shifts the patch to the beginning of the sample. This requires the use of \textit{zeor} to indicate the end of the patch.
\end{enumerate}

\begin{table}[!t]
\renewcommand{\arraystretch}{1.3}
\caption{\textbf{Data transformation.} Patch classification on the anomaly dataset. First row shows an invalid run as this setup performed a sample classification.}
\label{tab:accuracies_level2}
\centering
\begin{tabular}{|c|c|c|c|c|c|}
    \hline
	\multicolumn{3}{|c|}{\textbf{Setup}} & \multirow{2}{*}{\textbf{Train Acc.}} & \multirow{2}{*}{\textbf{Val Acc.}} & \multirow{2}{*}{\textbf{Test Acc.}} \\
	\textbf{zero} & \textbf{attach} & \textbf{notemp} & & & \\
	\hline
	\sout{False} & \sout{True} & \sout{False} & \sout{100.0} & \sout{98.54} & \sout{98.41} \\
	\hline
	True & False & False & 98.50 & 98.29 & 98.01 \\
	\hline
	True & True & False & 98.59 & 98.57 & 98.27 \\
	\hline
	True & False & True & 98.30 & 98.04 & 98.04 \\
	\hline
	True & True & True & 98.61 & 98.40 & 98.27 \\
	\hline
\end{tabular}
\end{table}

\begin{figure}[!t]
\centering
\includegraphics[width=0.9\linewidth]{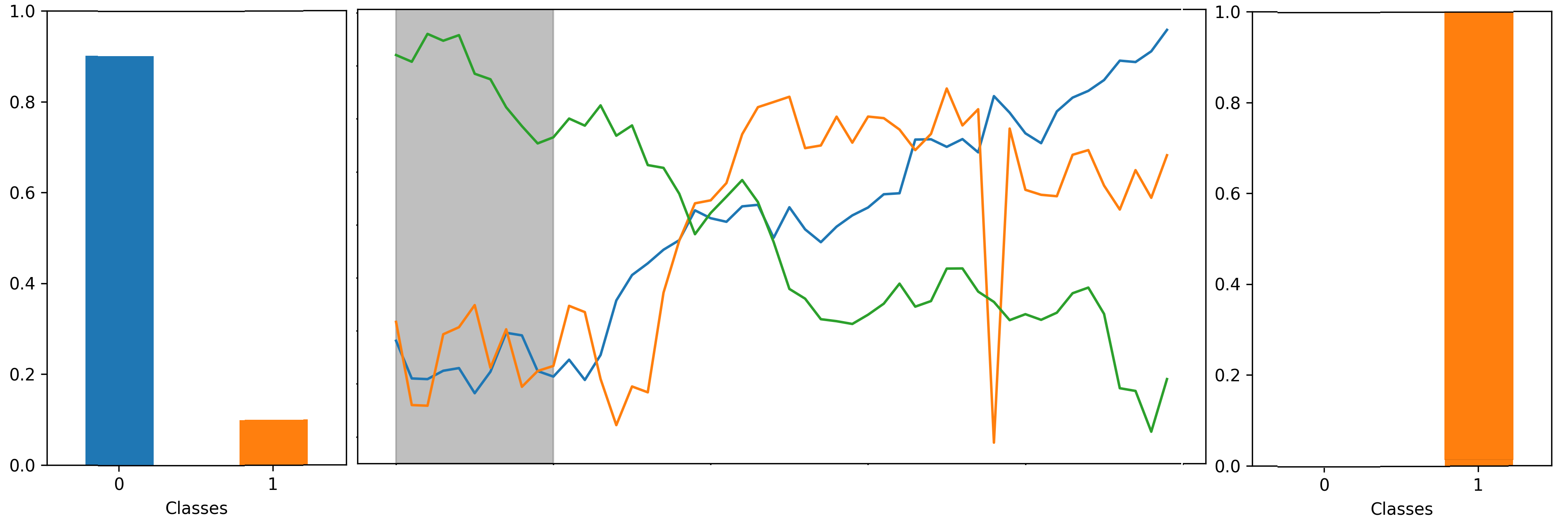}
\caption{\textbf{Patch classification.} Middle: Highlighted no anomaly part is classified. Left: Classification using \textit{zero}. Right: Classification not using \textit{zero}.}
\label{fig:anomaly_new_Patch_and_dist_0_0_comb}
\end{figure}

In Table~\ref{tab:accuracies_level2} we show different possible combinations. Surprisingly, the network was not able to exclude the data outside of the patches using the \textit{attach} transformation as shown in Figure~\ref{fig:anomaly_new_Patch_and_dist_0_0_comb} on the right side. The non-anomaly patch was classified as an anomaly and the saliency map has shown that the network took the peak into account. That means the network performed a sample classification instead of a patch classification. Using this information, we can conclude that the attachment of the channel is not enough to restrict the network. Therefore, setting the data not included in the patch to zero is mandatory to force a patch classification. In Figure~\ref{fig:anomaly_new_Patch_and_dist_0_0_comb} the result for the same patch using the \textit{zero} flag is shown on the left side. This time the model performed a patch classification. Using the other parameters can be of relevance when there is information available about the dataset. However, the performance increase for the \textit{Anomaly} dataset was limited due to the time and zero value independent task.

\subsection{Local and Global Patch-based Explanations}
Using the patch classifier it is possible to produce predictions for each patch. Based on the stride and length of each patch, this results in different overlapping explanations. As shown in Figure~\ref{fig:anomaly_new_Patch_and_dist_0_0_comb} it is possible to get detailed information about a patch and its classification. This patch explanation covers only a small piece of data and is much easier to understand due to the lower cognitive load required to capture the data~\cite{paas2003cognitive}. 

The combination of the patches results in a global sample explanation with fine-grained classification scores for every patch. Also, the height of the bars visualizes the confidence of the classifier. Lower bars correspond to patches that are not class-specific whereas higher values indicate a stronger relation to the picked class. 

\begin{figure}[!t]
\centering
\includegraphics[width=0.9\linewidth]{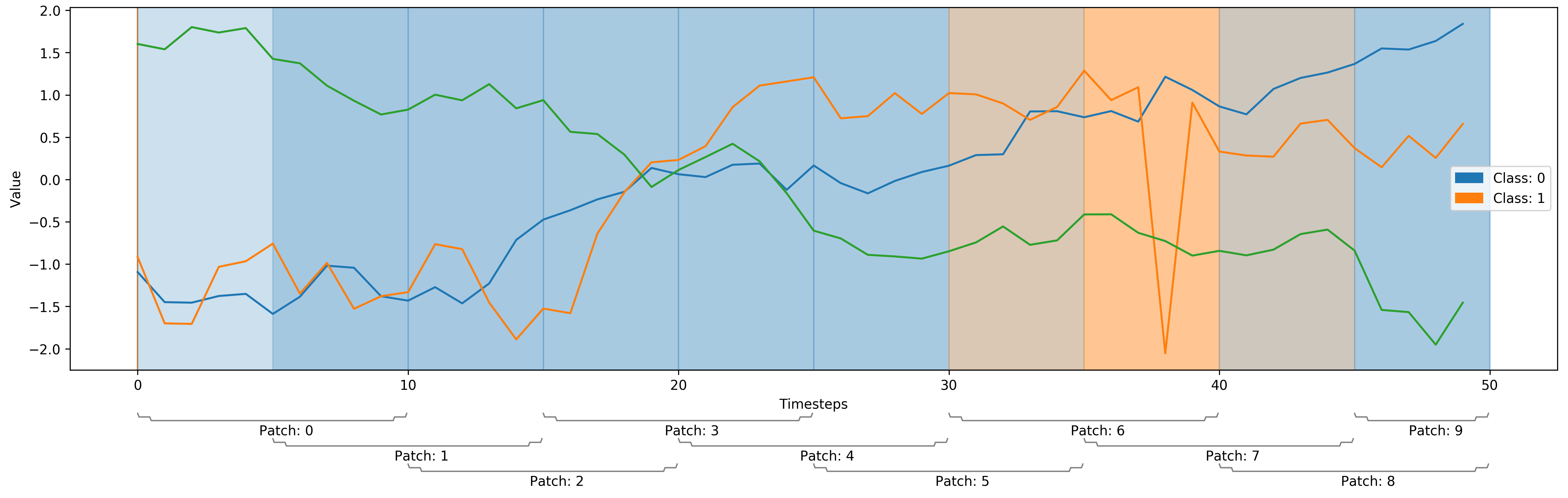}
\caption{\textbf{Explanation overlay.} Overlay of the patch classification. Gradient highlights confidence.}
\label{fig:anomaly_new_Class_overlay_0}
\end{figure}

In Figure~\ref{fig:anomaly_new_Class_overlay_0} we show the combination of the patch classifications. The gradient of the color further highlights the confidence. The blue color is assigned to non-anomaly patches and orange to anomaly patches. Ultimately, it highlights that the patch length for this sample could be even smaller. Precisely, some data classified as a non-anomaly by neighborhood patches is included in the anomaly patches. Using a setup with multiple patch sizes results in more precise localization of the anomaly. 

\begin{figure}[!t]
\centering
\subfloat[No anomaly patches]{\includegraphics[width=0.45\linewidth]{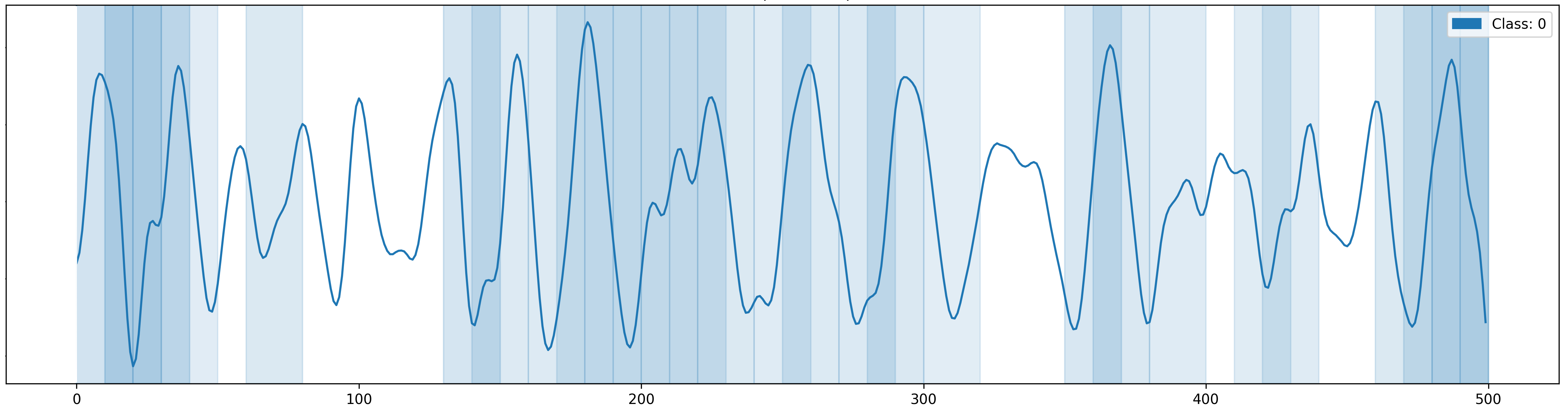}}
\label{fig:FordA_Class_overlay_0_c-0}
\subfloat[Anomaly patches]{\includegraphics[width=0.45\linewidth]{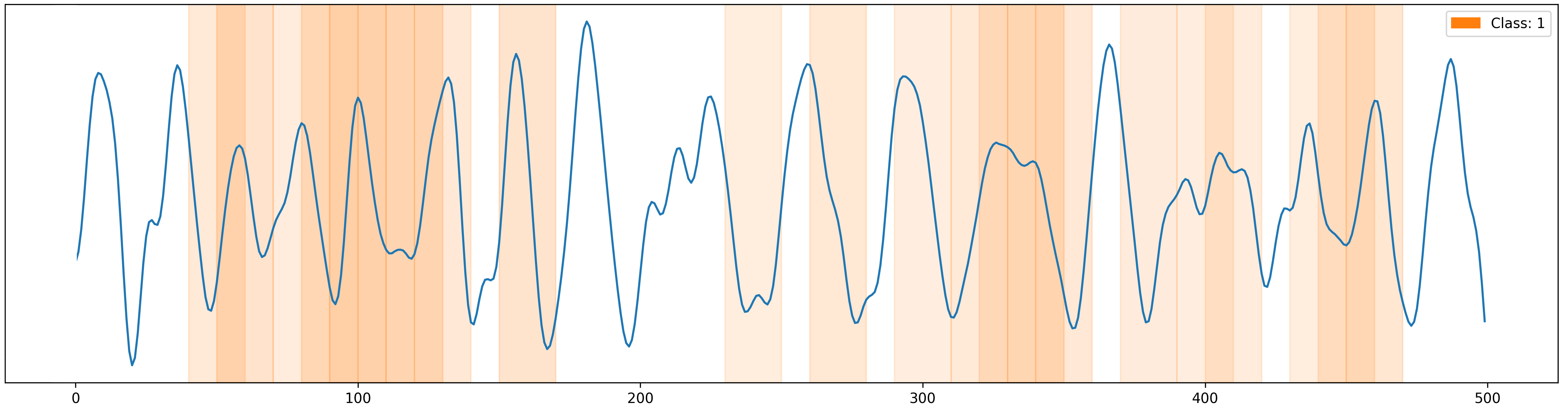}}
\label{fig:FordA_Class_overlay_0_c-1}

\caption{\textbf{FordA sequence explanation.} Class-wise patch prediction overlay using PatchX.}
\label{fig:FordA_Class_overlay_0}
\end{figure}

Besides, the overlay approach can provide visualizations restricted to a specific set of classes .g. an explanation of the \textit{FordA} dataset consists of two classes. This dataset covers an anomaly detection task. In contrast to the previous results using the \textit{Anomaly} dataset, the anomalies are not restricted to a single location. However, it is possible to highlight the class-specific regions, as shown in Figure~\ref{fig:FordA_Class_overlay_0}. Especially, when the data covers a large number of time-steps and data unrelated to the class of interest this visualization enables easy filtering. Precisely, the explanation shown in Figure~\ref{fig:FordA_Class_overlay_0} highlights small parts there were classified independently as an anomaly or no anomaly enabling a piece-wise inspection of these pieces.

\subsection{Global Patch Confidence}
As the local patch classifier predicts a label for each patch it is important to differ between the following cases. Patches that have high value are likely to be class-specific as they only appear in a small number of classes. Patches that have a low value are likely to be class unrelated as they across the classes. However, there exist patches that are shared between classes and have medium values. 

\begin{figure}[!t]
\centering
\subfloat[]{\includegraphics[width=0.225\linewidth]{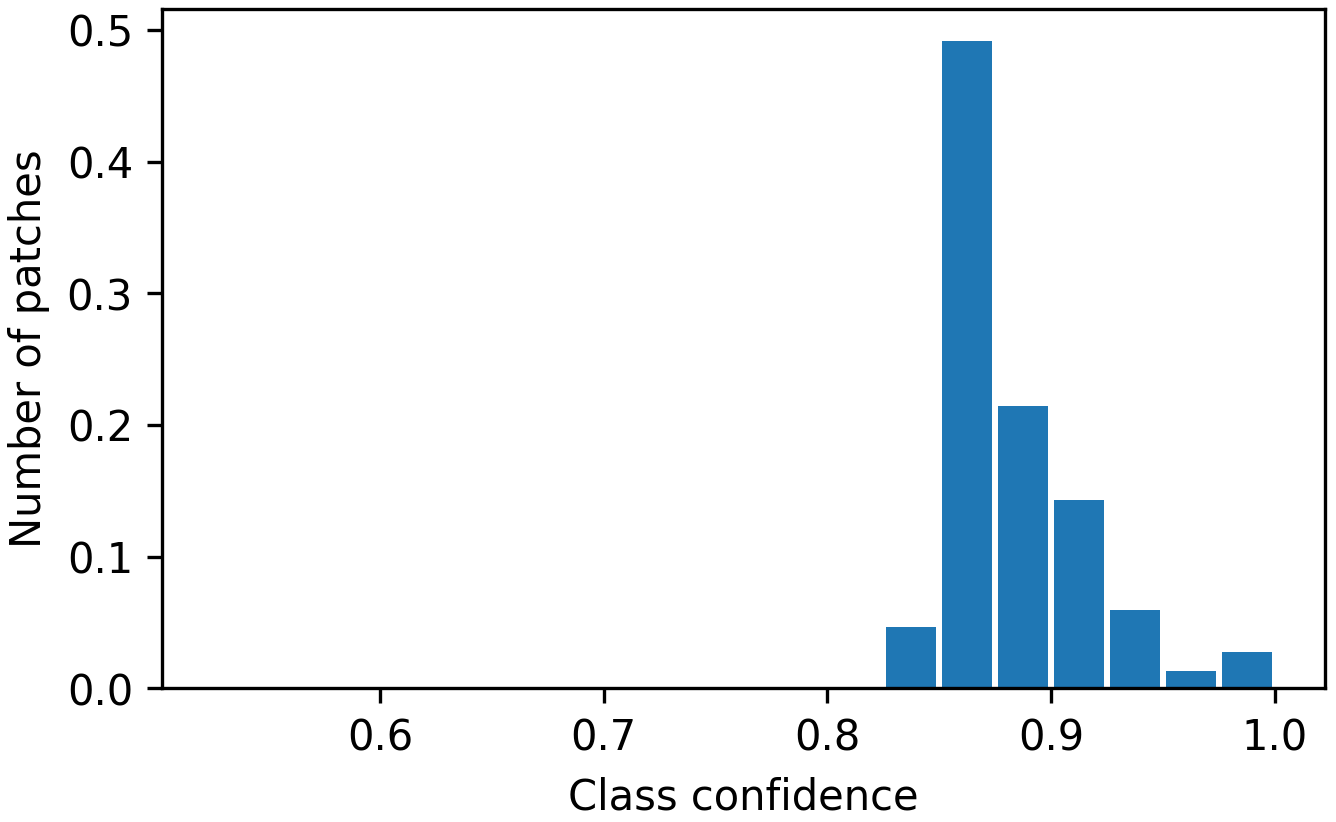}}
\label{fig:anomaly_new_Patch_relevance}
\subfloat[]{\includegraphics[width=0.225\linewidth]{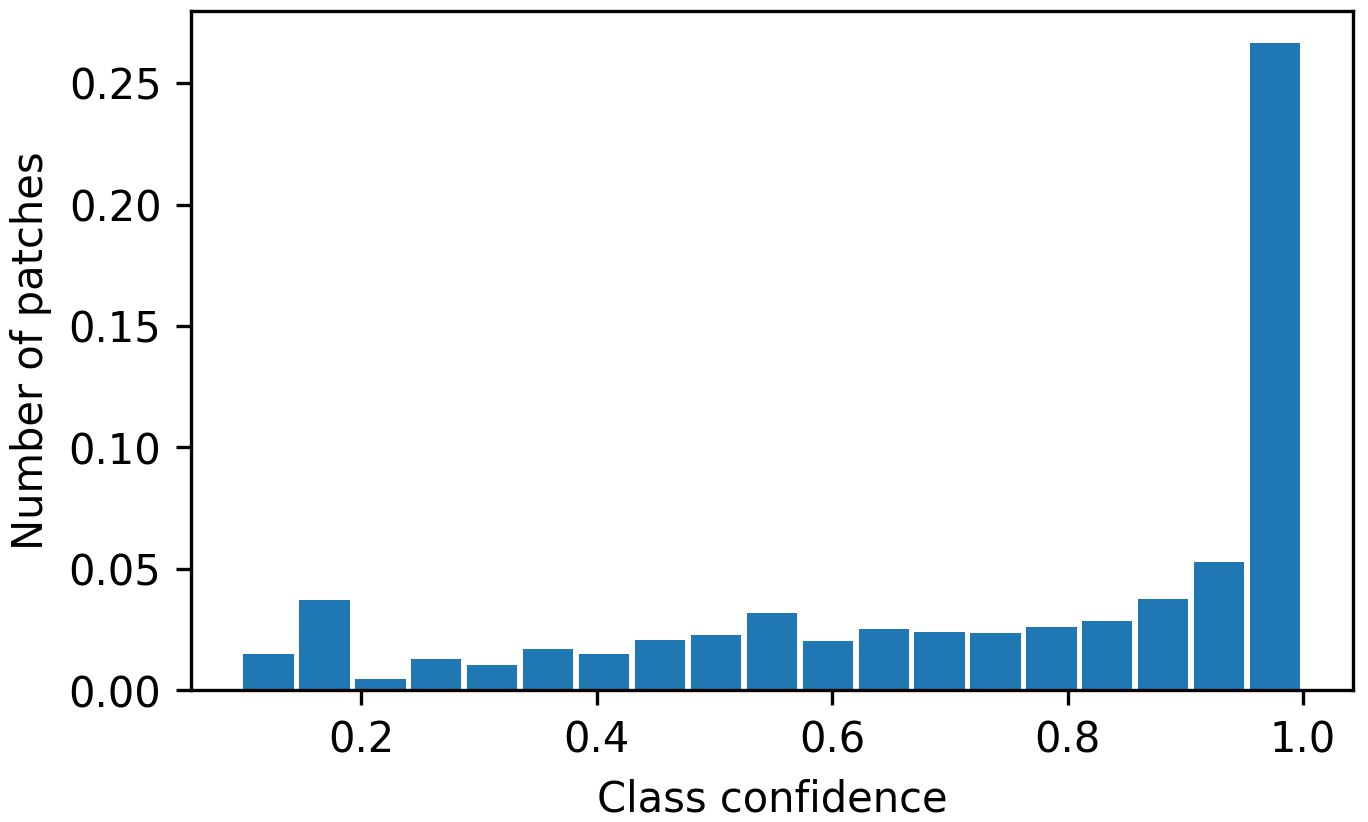}}
\label{fig:character_trajectories_Patch_relevance}
\subfloat[]{\includegraphics[width=0.225\linewidth]{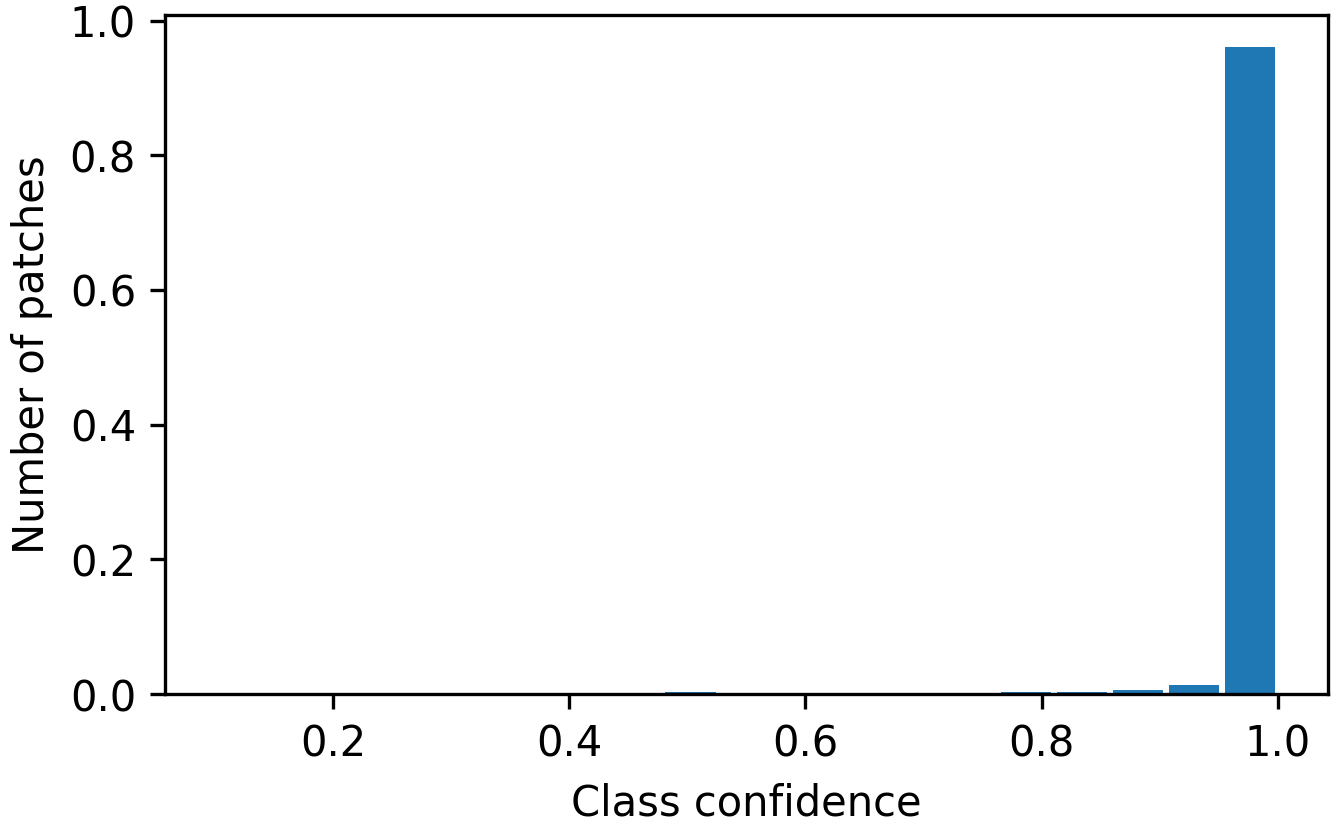}}
\label{fig:daily_and_sport_activites_Patch_relevance}
\subfloat[]{\includegraphics[width=0.225\linewidth]{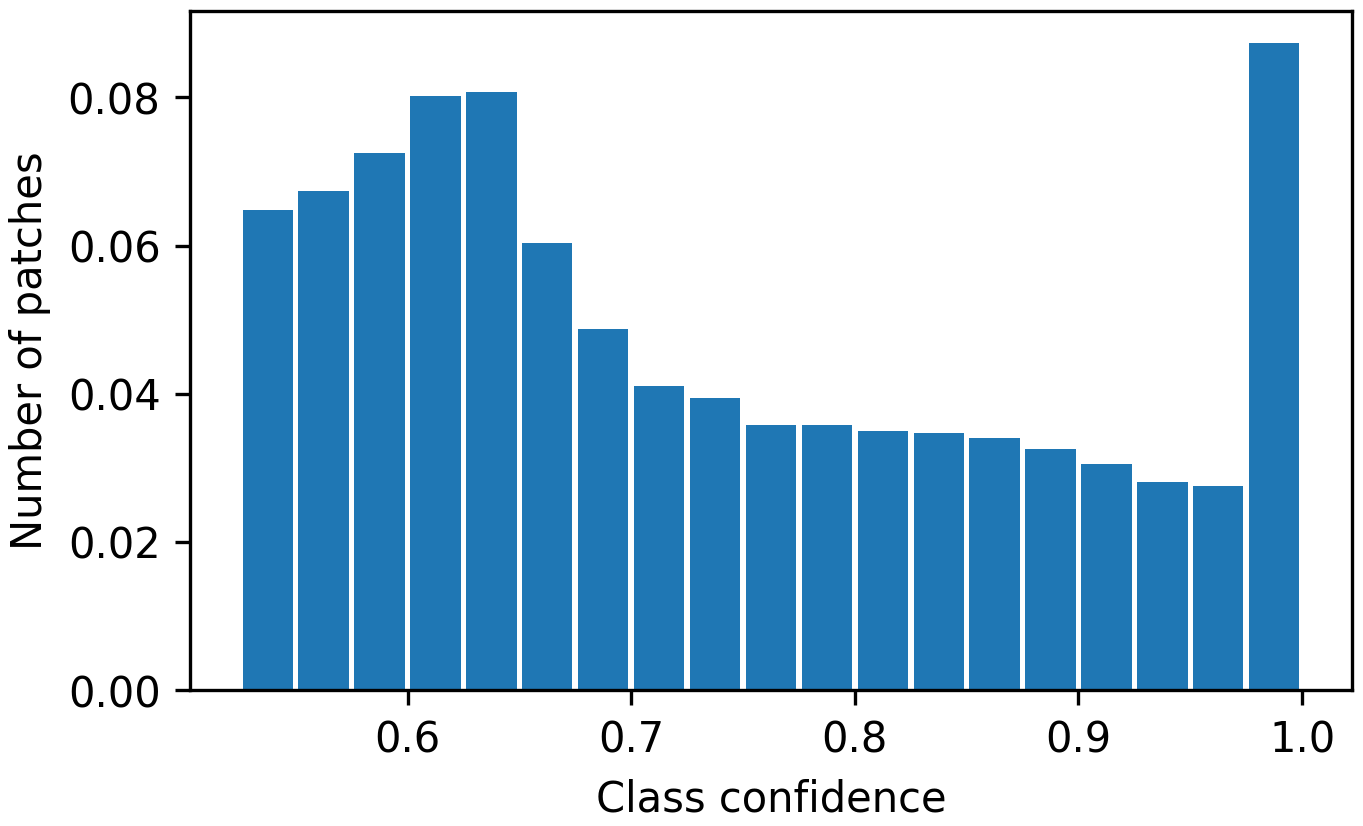}}
\label{fig:FordA_Patch_relevance}

\caption{\textbf{Patch confidence.} Y-axis: number of patches. X-axis: Soft-max. Left to right: Anomaly dataset, Character trajectories, DailyAndSportActivities, and FordA.}
\label{fig:Patch_relevance}
\end{figure}

Using the scores for each patch it is possible to get global insights about the task. In Figure~\ref{fig:Patch_relevance} the patch confidence over the dataset is visualized as a histogram. The confidence of each patch is calculated using the value of their softmax prediction to highlight their uniqueness. The patterns of the \textit{DailyAndSportActivities} are unique to each class highlighted by the high values. The same holds for the \textit{character trajectories} dataset. However, some of the patterns are shared across the classes. Intuitively, when drawing a character some parts of the drawing are not unique to a character. E.g. the classes 'e' and 'o' only differ in the first and last part of the signal. In contrast to the previously mentioned datasets, the \textit{anomaly} dataset shows different behavior as it covers only point anomalies the number of patches that appear exclusively in one class is smaller. The \textit{FordA} dataset shows a combination of the previously mentioned behaviors as it contains class-specific, unrelated, and shared patterns.

\subsection{Class Boundary Evaluation}
\begin{figure}[!t]
\centering
\includegraphics[width=0.6\linewidth]{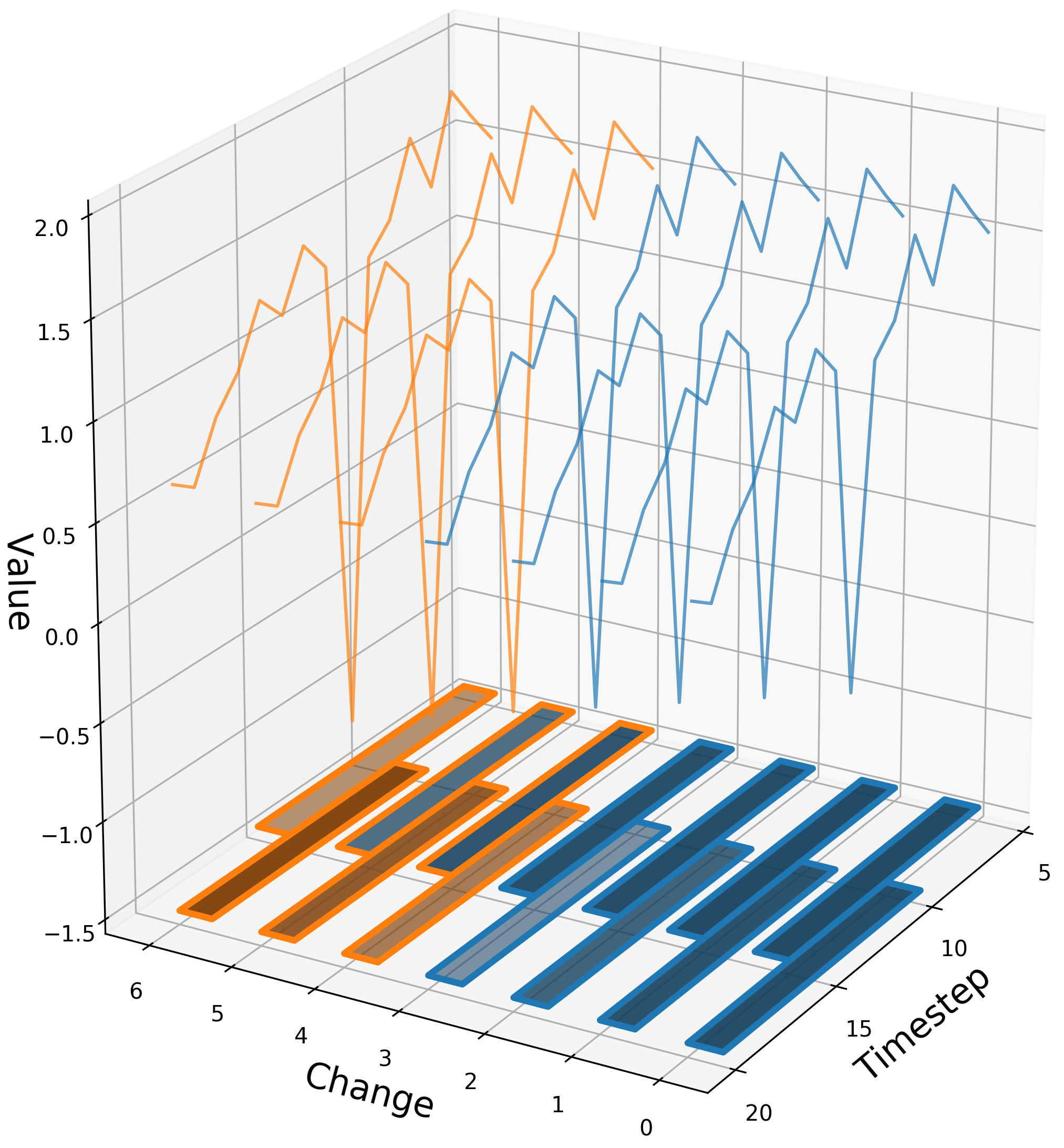}
\caption{\textbf{Class smoothing.} Line color: ground-truth. Bars border color: prediction. Bar color and the gradient: Patch prediction.}
\label{fig:anomaly_new_Class_Smoothing}
\end{figure}

The patch prediction and confidence scores can be used to explore class boundaries. We show that they help to identify the source of a class shift. In Figure~\ref{fig:anomaly_new_Class_Smoothing} a peak is visualized. The initial series (blue line) without any change has a non-anomaly label. When increasing the peak value, the label switches after the third change to an anomaly (orange line). Mathematically, the labels of this dataset are computed based on the mean and std within the signal. Therefore, it is possible to create a series of the same signal while slightly increasing the peak to shift the class. However, if we want to understand if the classifier learned the class boundary correctly, we have to take a look at the change of the patch and overall prediction. The border color of the bars shows the sample classification, whereas the bar color the patch prediction. We used two patches for this sample. The results provide evidence that the boundary is learned correctly, as the border color of the bars and the line color match. However, the patch covering time-steps 10 to 20 changes first, whereas the prediction of the patch that only includes a part of the peak changes much slower. 

\begin{figure}[!t]
\centering
\subfloat[Label: 'o'. Prediction: 'e']{\includegraphics[width=0.45\linewidth]{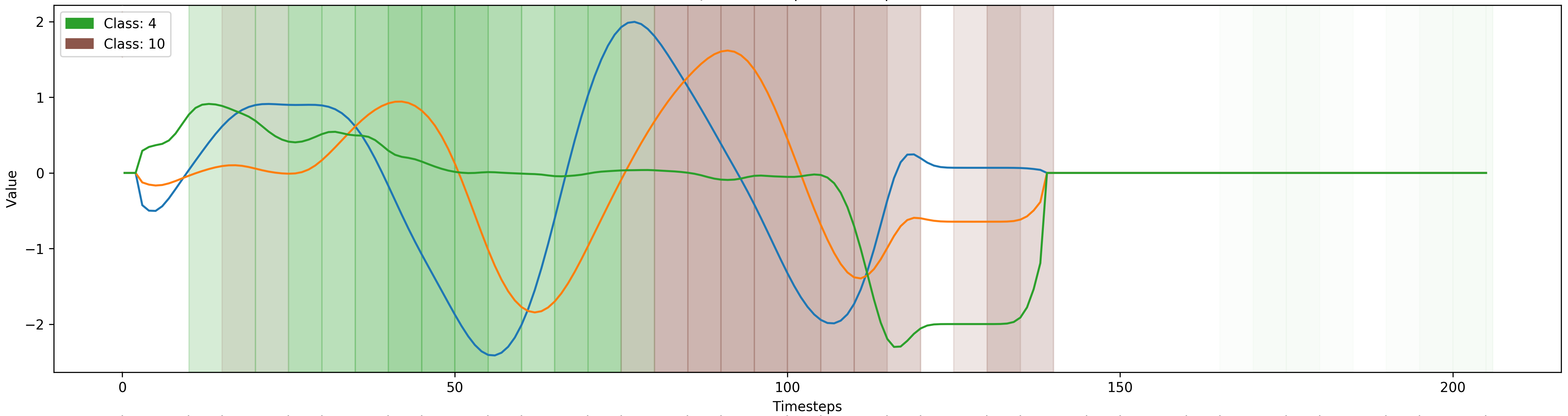}}
\label{fig:character_trajectories_Class_overlay_789_c-4-10}
\subfloat[Label: 'w'. Prediction: 'u']{\includegraphics[width=0.45\linewidth]{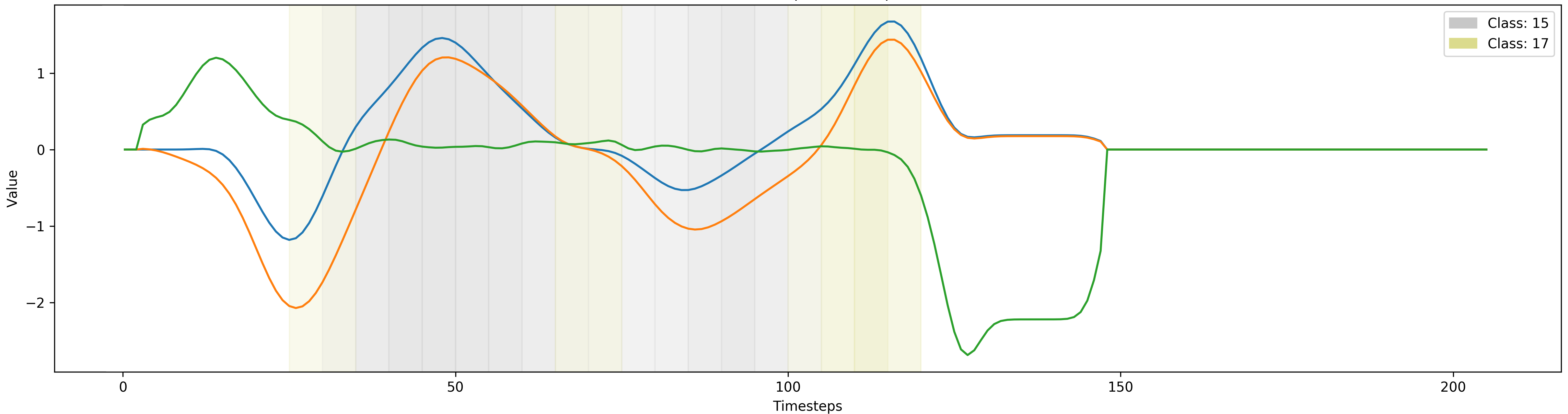}}
\label{fig:character_trajectories_Class_overlay_1195_c-15-17}

\subfloat[Pred. as: 'o'.]{\includegraphics[width=0.225\linewidth]{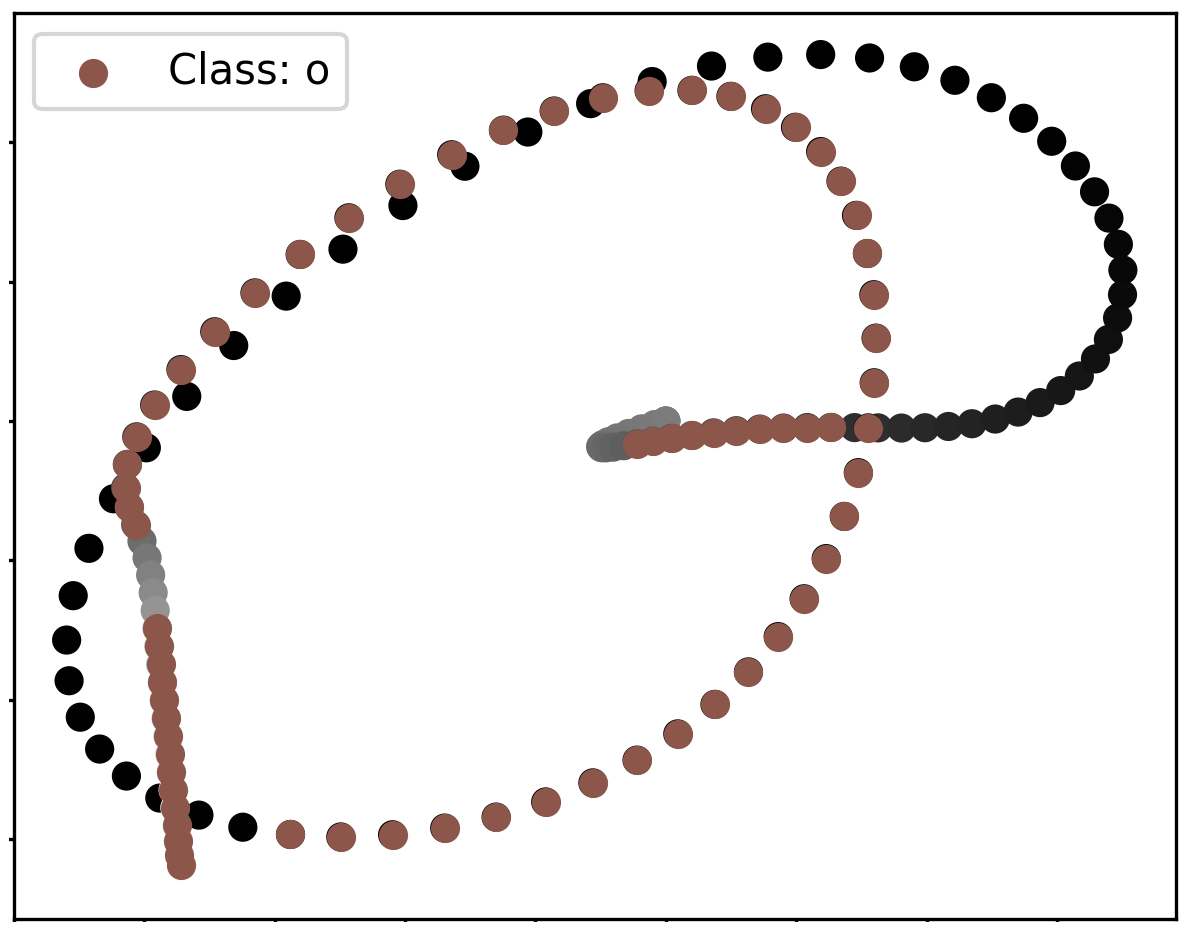}}
\label{fig:character_trajectories_Class_overlay_char_789_c-10}
\subfloat[Pred. as 'e'.]{\includegraphics[width=0.225\linewidth]{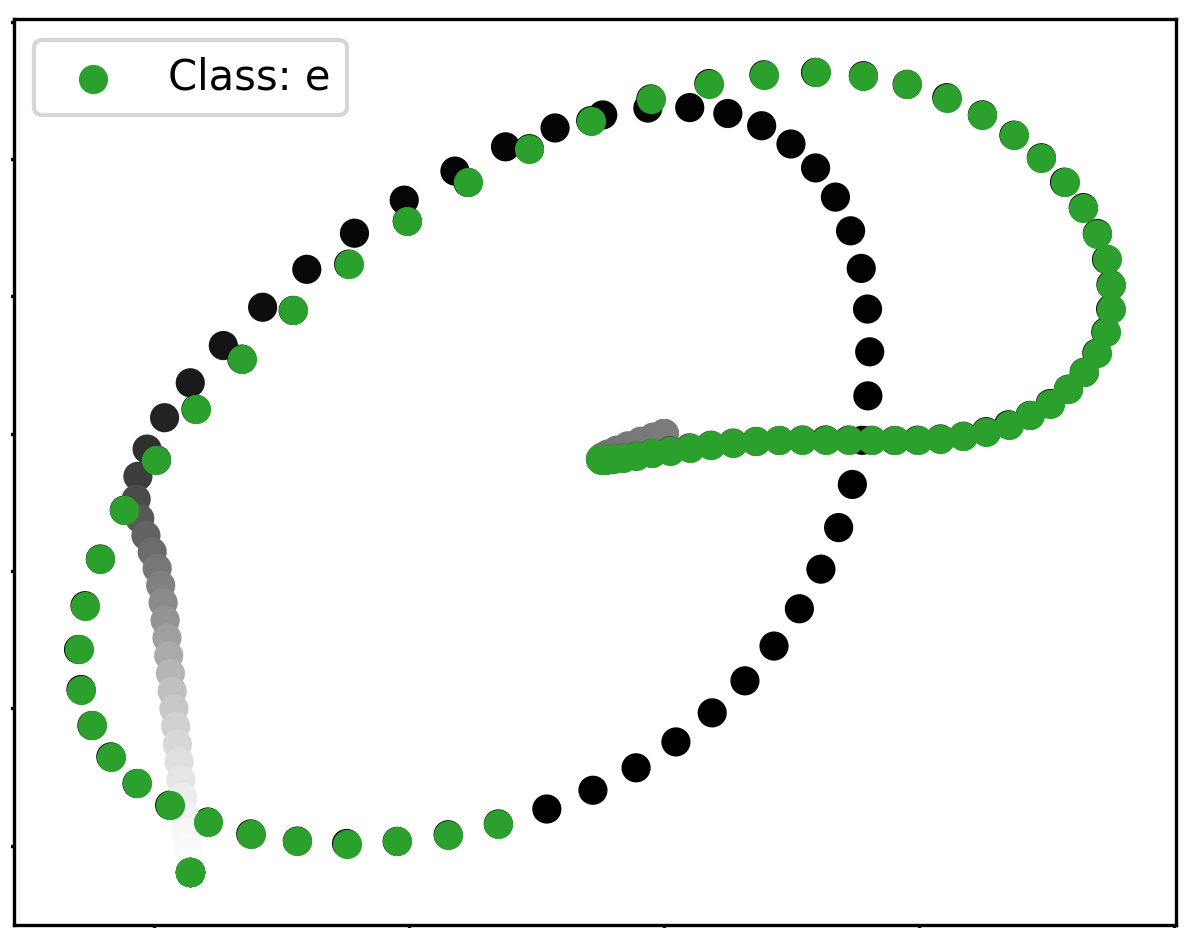}}
\label{fig:character_trajectories_Class_overlay_char_789_c-4}
\subfloat[Pred. as 'u'.]{\includegraphics[width=0.225\linewidth]{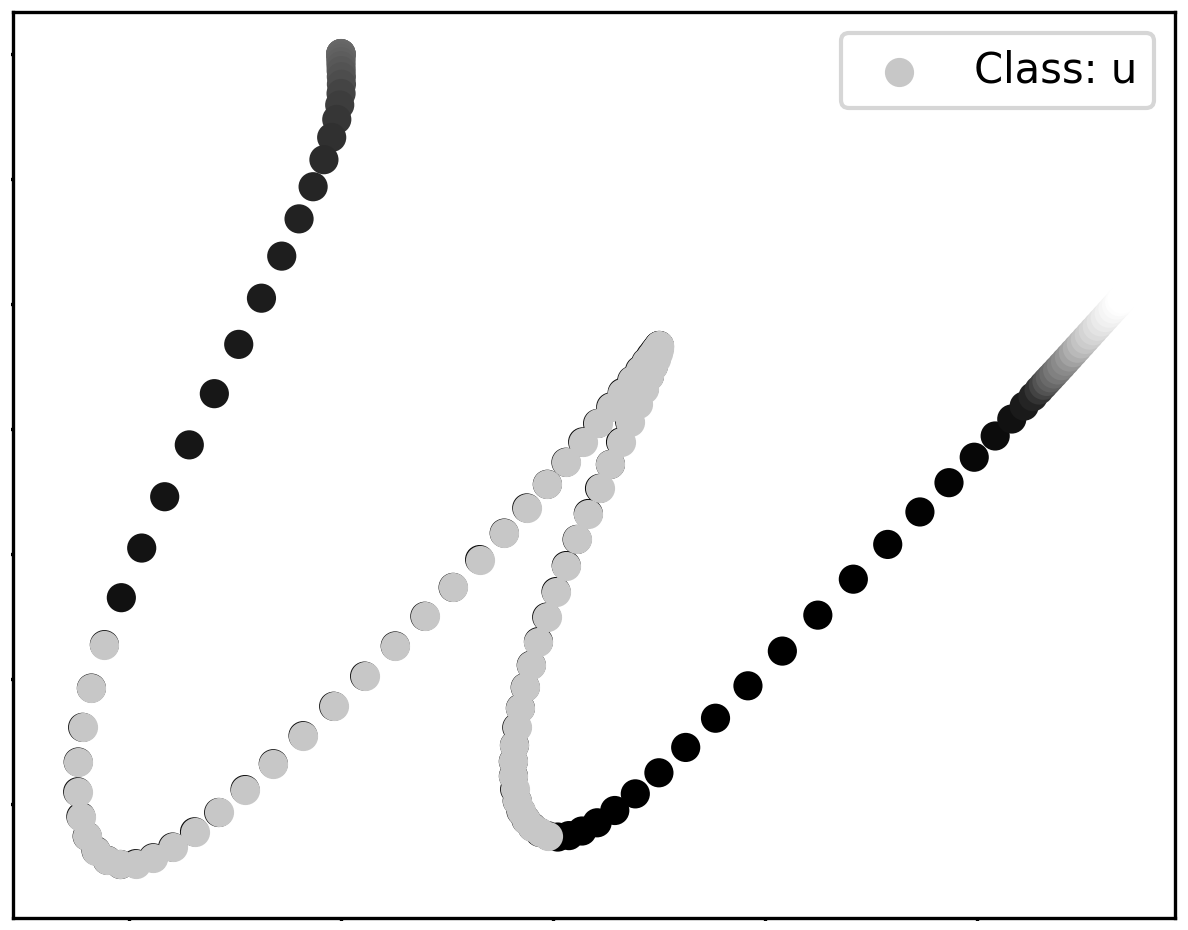}}
\label{fig:character_trajectories_Class_overlay_char_1195_c-15}
\subfloat[Pred. as 'w'.]{\includegraphics[width=0.225\linewidth]{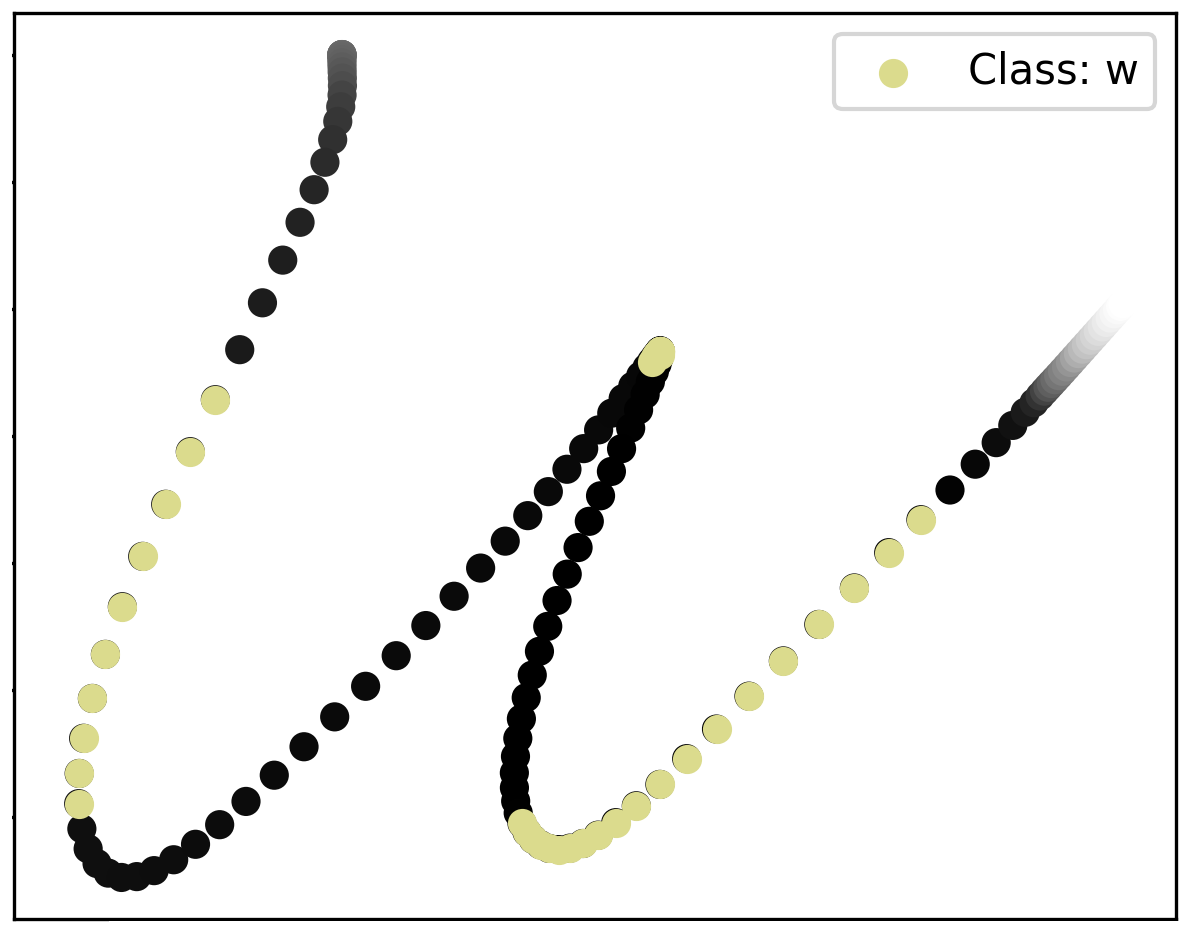}}
\label{fig:character_trajectories_Class_overlay_char_1195_c-17}

\caption{\textbf{Mislabel explanation.} First row: Overlay of two mislabeled characters. Second row: Correct classified patches. Third row: Misclassified patches.}
\label{fig:character_trajectories_misclassification}
\end{figure}

Also, samples sometimes are mislabeled, because they are close to a class boundary. Then, the question arises whether the approach or the label is correct. In Figure~\ref{fig:character_trajectories_misclassification} the results of two misclassified characters are shown. With the help of the patch-wise prediction, it is possible to directly understand the classification. We projected back the time-series to the 2d drawing of the character and used the overlay to highlight the predicted classes for the patches.

\subsection{Comparison with State-of-the-art Approaches}
\begin{figure}[!t]
\centering
\subfloat[PatchX]{\includegraphics[width=0.3\linewidth]{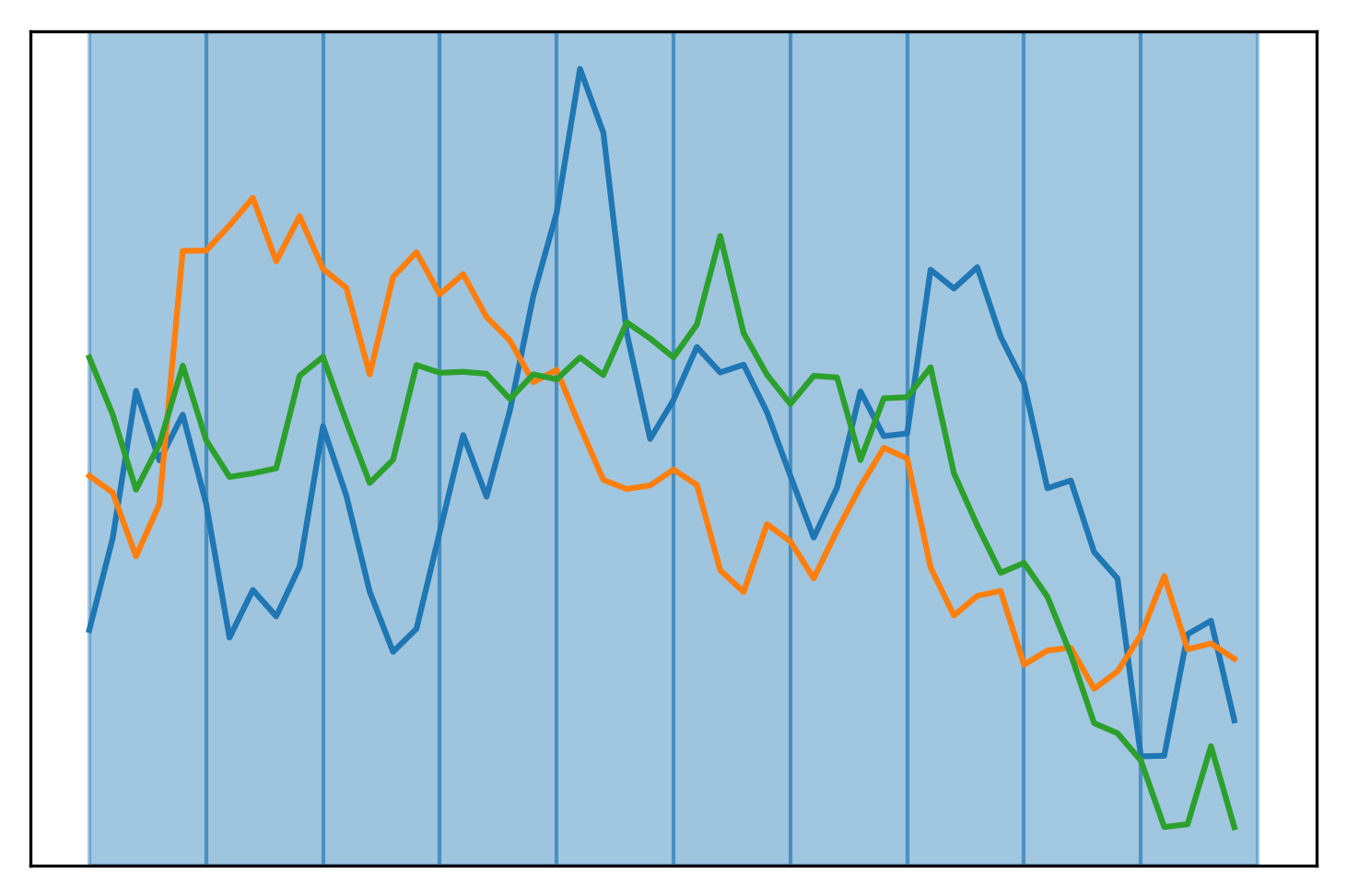}}
\label{fig:sota_patchx_1}
\subfloat[SHAP]{\includegraphics[width=0.3\linewidth]{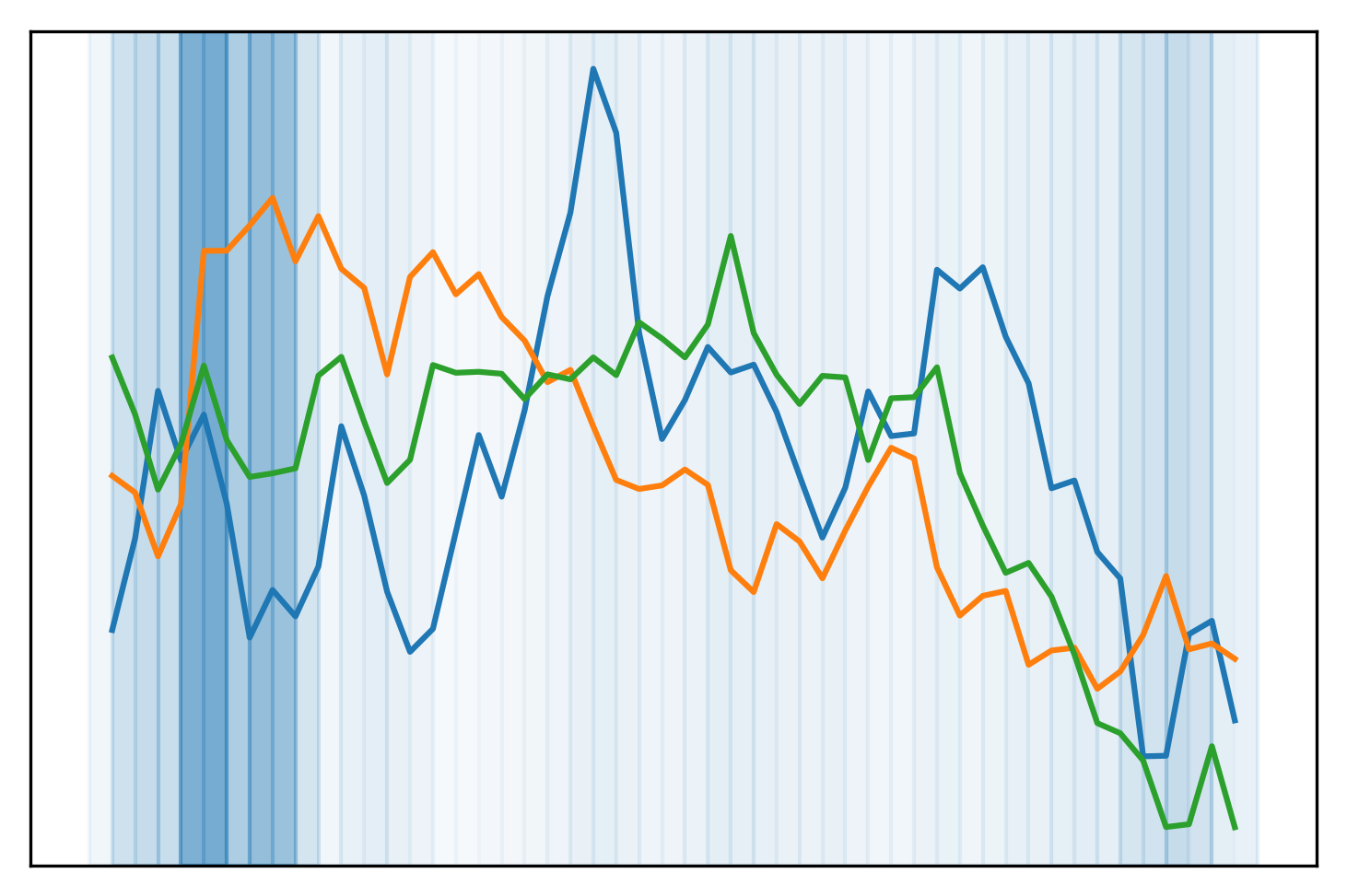}}
\label{fig:sota_shap_1}
\subfloat[LIME]{\includegraphics[width=0.3\linewidth]{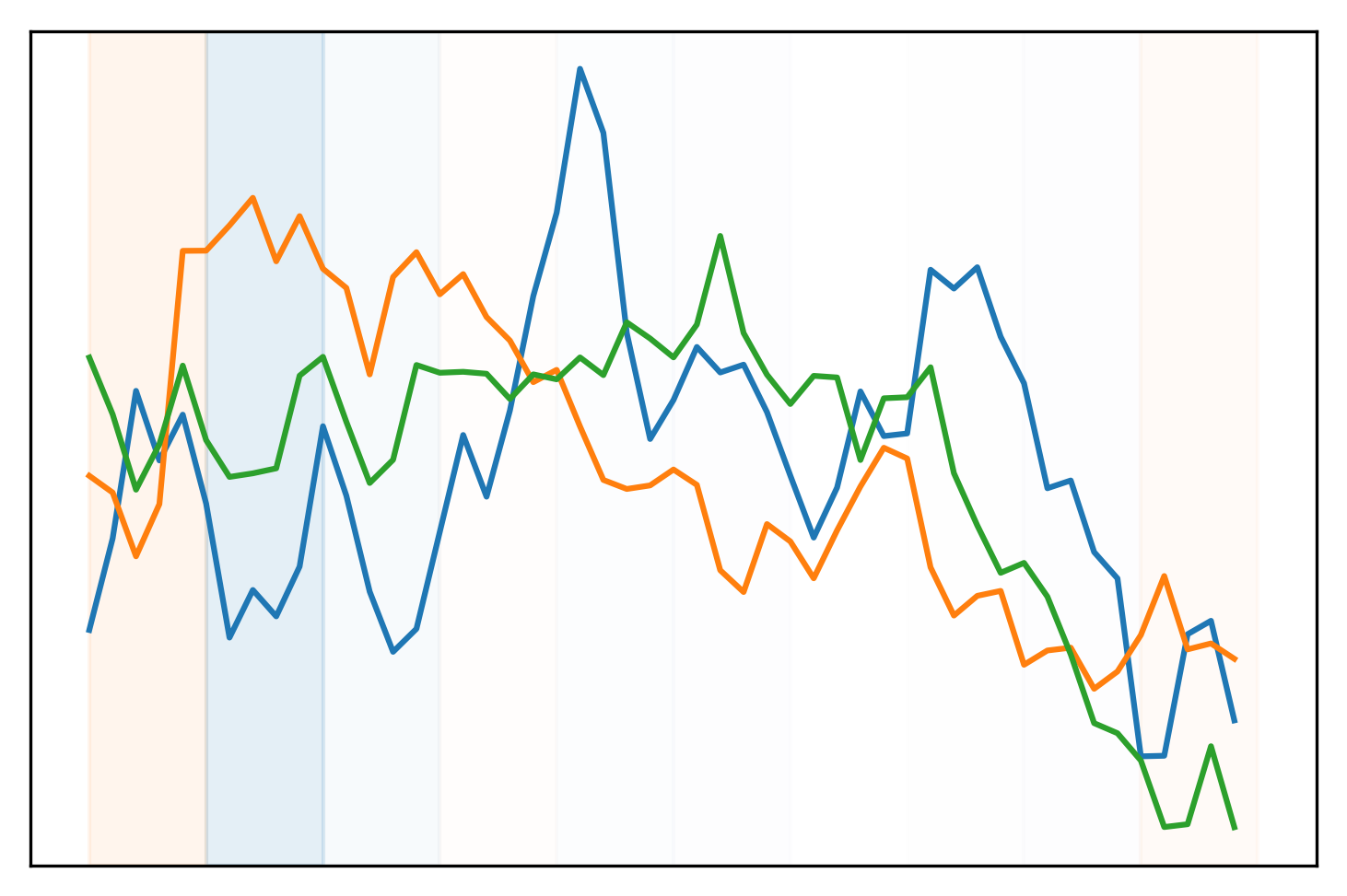}}
\label{fig:sota_lime_1}

\subfloat[PatchX]{\includegraphics[width=0.3\linewidth]{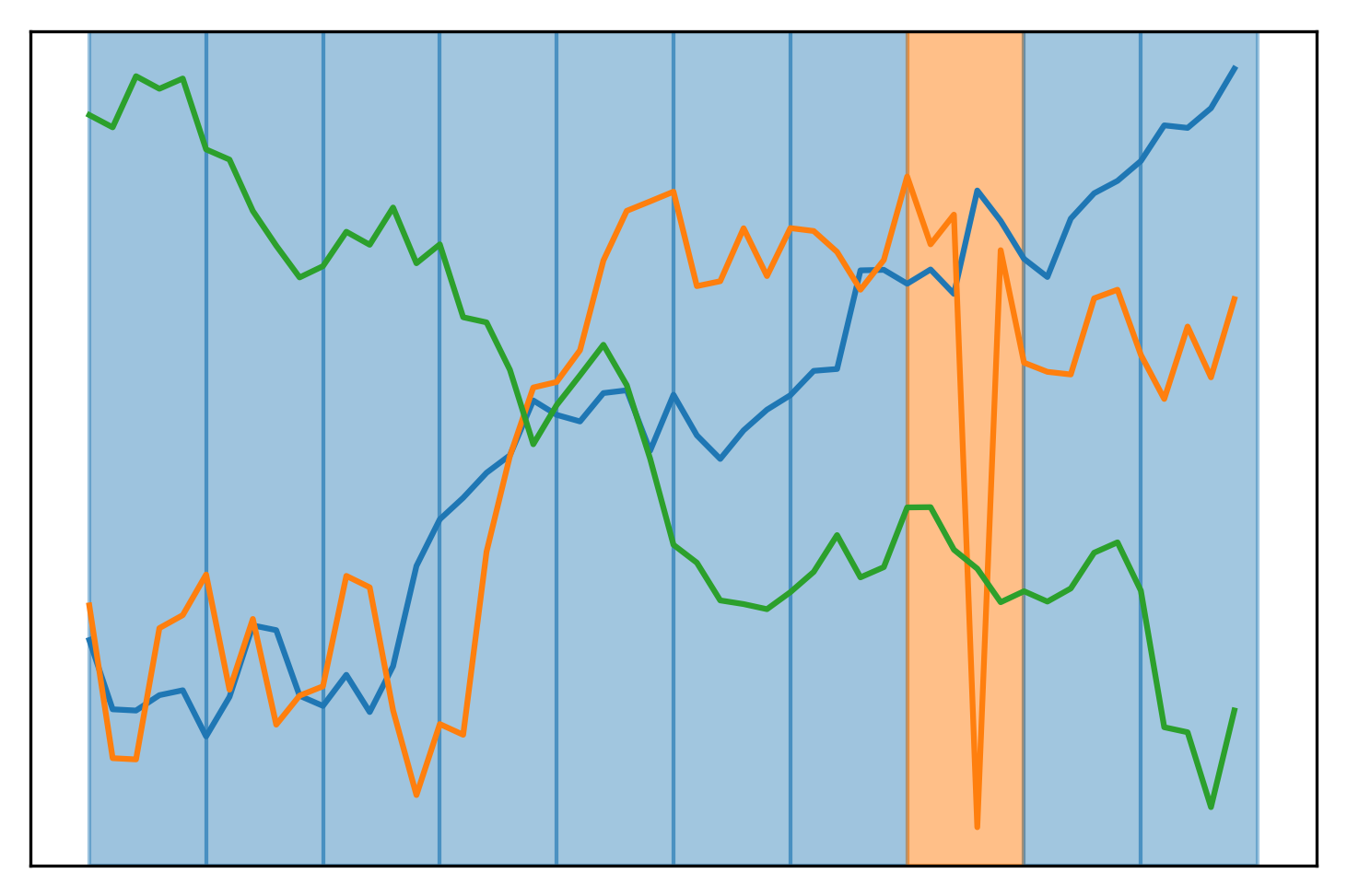}}
\label{fig:sota_patchx_0}
\subfloat[SHAP]{\includegraphics[width=0.3\linewidth]{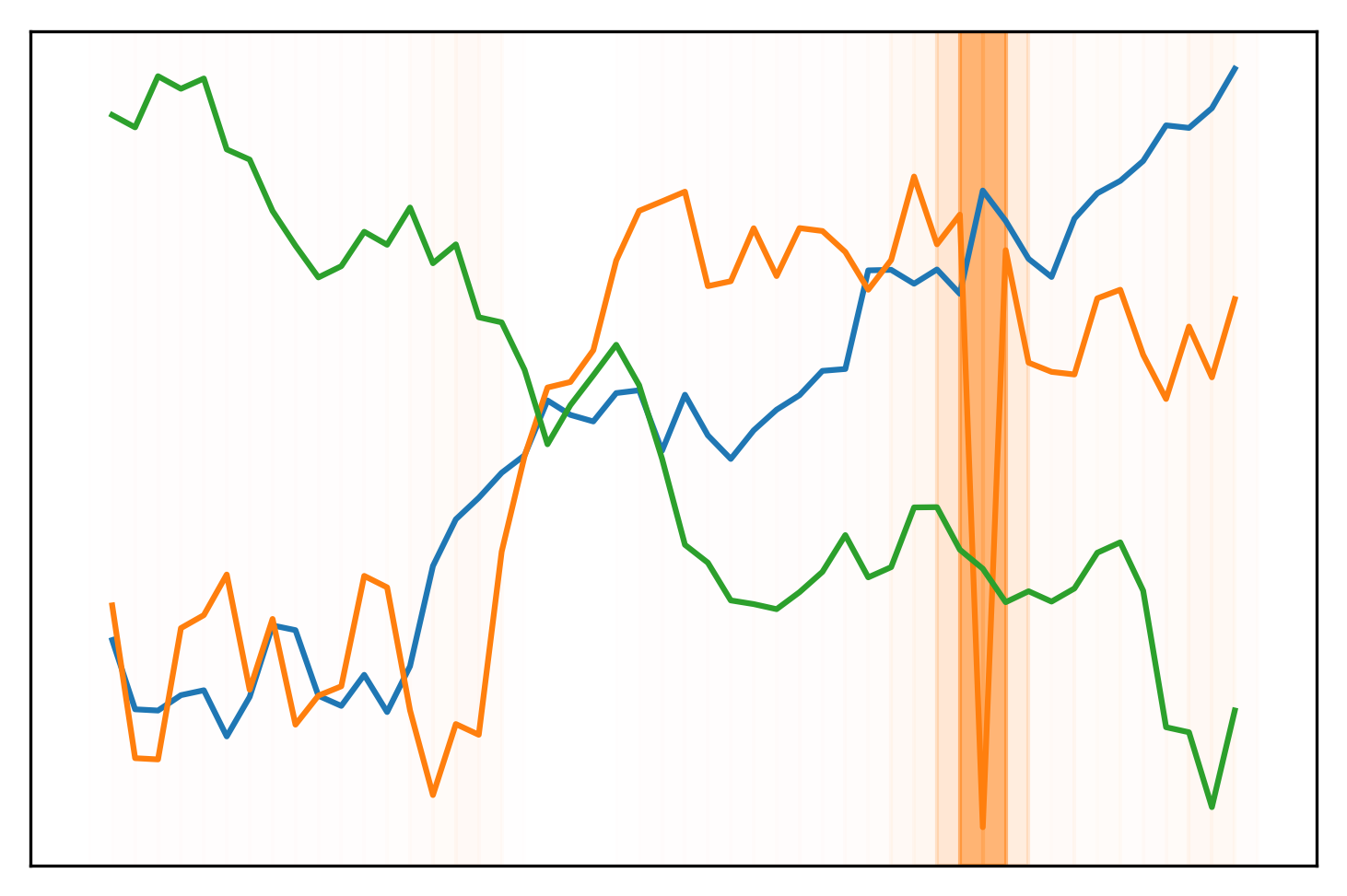}}
\label{fig:sota_shap_0}
\subfloat[LIME]{\includegraphics[width=0.3\linewidth]{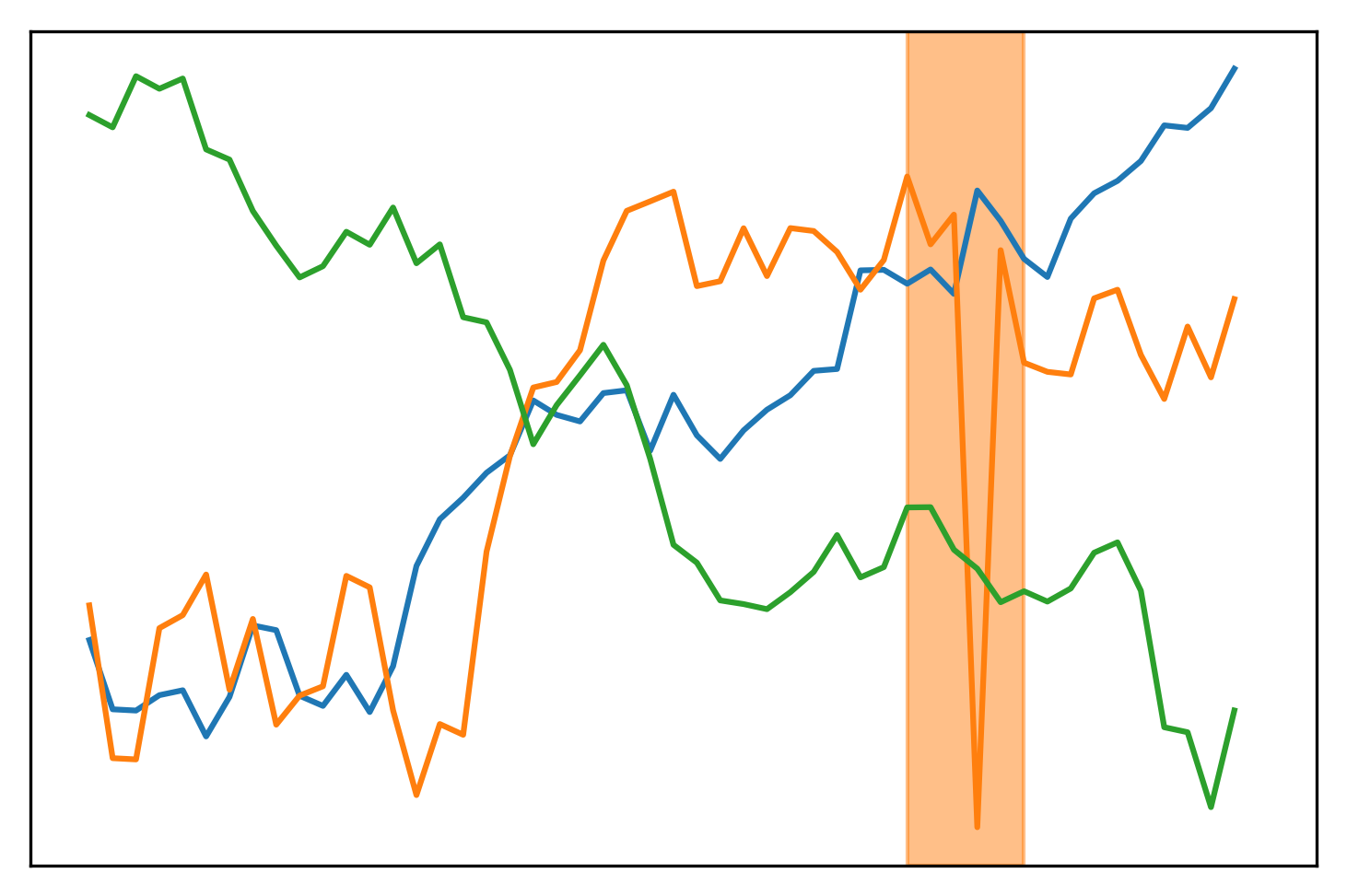}}
\label{fig:sota_lime_0}

\caption{\textbf{Comparison SotA.} First row no anomaly, and second row anomaly sample. Orange: Anomaly. Blue: No anomaly. White: not classified.}
\label{fig:comparison_sota}
\end{figure}

This section provides a comparison of our method with two well-known state-of-the-art approaches namely LIME~\cite{ribeiro2016should} and SHAP~\cite{lundberg2017unified}. As both methods use the blackbox model to perform a model-agnostic explanation the accuracy can be found in Table~\ref{tab:accuracies_comparison} using the blackbox model. It has to be mentioned that both approaches are designed mainly for tabular and image data. However, compared to the other approaches mentioned in Table~\ref{tab:related} they are the most suitable and best-performing choice for a comparison. Figure~\ref{fig:comparison_sota} shows two samples of the anomaly dataset. Starting with the second row, all methods were able to precisely locate the peak within the signal. However, SHAP and LIME only provide information about the part relevant to the prediction. In contrast, our approach provides information for every patch. The first row shows a sample with any significant peak. However, the explanation of SHAP and LIME is not as intuitive as ours. 

It has to be mentioned that SHAP provides additional information about the relevant channel and it is possible to combine the approaches. SHAP can be used on our fine-grained patch prediction model to provide additional insights about the neural network to enhance the explanation further. The same holds for every other model agnostic method mentioned in the related work. A comparison of the other methods is rather difficult as the focus on other aspects e.g. global prototypes or data compression.

\section{Conclusion}
We have shown that our hybrid approach can produce interpretable results for different time-series classification tasks. Our utilization of neural networks and traditional machine learning approaches provides local and global instance-based explanations. The approach improves the interpretation of mislabels, class boundaries, and sample explanation. Furthermore, our hybrid approach covers different levels of explanation focusing on both low and high-level patterns. Finally, our results emphasize that our hybrid approach builds a bridge between the interpretable traditional machine learning algorithms and neural networks. It combines the scalability, performance, and interpretability advantages of both worlds. In future work, we will investigate aspects concerning the automatic parameter tuning for the stride and patch size selection and a more advanced training process taking into account the relevance of a patch.

\section*{Acknowledgements}
This work was supported by the BMBF projects DeFuseNN (Grant 01IW17002) and the ExplAINN (BMBF Grant 01IS19074). We thank all members of the Deep Learning Competence Center at the DFKI for their comments and support.

\bibliographystyle{IEEEtran}
\bibliography{bibliography}

\end{document}